\documentclass[a4paper, 11pt]{article}

\usepackage[utf8]{inputenc}
\usepackage[T1]{fontenc}  
\usepackage[english]{babel}
\usepackage{hyperref, url, amsmath, amssymb}
\usepackage[ruled]{algorithm2e}
\usepackage{authblk}
\usepackage[margin=2cm]{geometry}

\usepackage{booktabs}
\usepackage{lscape}
\usepackage{cite}
\usepackage{mathtools,amssymb,amsmath}
\usepackage{enumerate}
\usepackage{varioref}
\usepackage{amsmath,mathpazo}
\usepackage{cleveref}
\usepackage{xcolor}
\usepackage{algpseudocode}
\usepackage{listings}
\usepackage{csquotes}
\usepackage{url}
\usepackage{subcaption}
\usepackage{multirow}
\usepackage{float}

\newcommand{\dir}{\mbox{Dir}}
\newcommand{\ddim}{\mbox{dim}}
\newcommand{\conv}{\mbox{Conv}}
\newcommand{\convT}{\mbox{ConvT}}
\newcommand{\lin}{\mbox{Linear}}
\newcommand{\resh}{\mbox{Reshape}}

\newcommand{\best}{\mbox{{\scriptsize best}}}
\providecommand{\keywords}[1]
{
  \small	
  \textbf{\textit{Keywords---}} #1
}

\title{Simplex Autoencoders}

\author[1,2]{Aymene Mohammed Bouayed\thanks{Corresponding author: \texttt{Aymene.Bouayed@ens.fr}}}
\author[1]{David Naccache\thanks{\texttt{david.naccache@ens.fr}}}
\affil[1]{DIENS, ENS, PSL University, Paris, France}
\affil[2]{Be-Ys Research, France}
\date{}

\begin{document}

\maketitle

\begin{abstract}
Synthetic data generation is increasingly important due to privacy concerns. While Autoencoder-based approaches have been widely used for this purpose, sampling from their latent spaces can be challenging. Mixture models are currently the most efficient way to sample from these spaces. In this work, we propose a new approach that models the latent space of an Autoencoder as a simplex, allowing for a novel heuristic for determining the number of components in the mixture model. This heuristic is independent of the number of classes and produces comparable results. We also introduce a sampling method based on probability mass functions, taking advantage of the compactness of the latent space. We evaluate our approaches on a synthetic dataset and demonstrate their performance on three benchmark datasets: MNIST, CIFAR-10, and Celeba. Our approach achieves an image generation FID of $4.29$, $13.55$, and $11.90$ on the MNIST, CIFAR-10, and Celeba datasets, respectively. The best AE FID results to date on those datasets are respectively $6.3$, $85.3$ and $35.6$ we hence substantially improve those figures (the lower is the FID the better). However, AEs are not the best performing algorithms on the concerned datasets and all FID records are currently held by GANs. While we do not perform better than GANs on CIFAR and Celeba we do manage to squeeze-out a non-negligible improvement (of 0.21) over the current GAN-held record for the MNIST dataset.
\end{abstract}

\noindent \keywords{Autoencoder, Synthetic image generation, Latent space sampling, Probability mass function.}

\section{Introduction}

Living in an era where a lot of data is available has many benefits for training artificial neural networks to perform various tasks and achieve good performance. However, with the growing concern about privacy and the implementation of GDPR regulations, the generation of synthetic datasets that mimic the distribution of real-world datasets has become increasingly important. Many works have been done in this area, with most approaches relying on two main neural network architectures: autoencoders \cite{Goodfellow-et-al-2016} and generative adversarial networks \cite{GAN}.

Models based on the GAN architecture train two neural networks - a generator and a discriminator - in an adversarial fashion. The generator attempts to transform a given distribution to match the distribution of real-world examples, while the discriminator tries to differentiate between synthetic and real-world data. GAN-based models can produce high-quality, high-fidelity data, particularly images \cite{style_gan_2}. However, due to the adversarial nature of their training, which involves finding a Nash equilibrium, GANs can be difficult to train. In many cases, GAN models without careful hyperparameter tuning can collapse and produce identical data points \cite{collapse_mode_GAN}.

The other family of synthetic data generation is based on autoencoders. Autoencoders are artificial neural networks that can learn to efficiently encode unlabeled data and belong to a broader category of unsupervised learning algorithms. Autoencoders consist of two neural networks - an encoder and a decoder. The encoder maps the input data to a lower-dimensional latent space, while the decoder reconstructs the input data based on the encoding generated by the encoder. This type of model progressively validates and refines its model weights by using the model to regenerate inputs from the current encoding and minimize the reconstruction loss. The core of the autoencoding approach involves training the model to discard noise - insignificant data that is of no or little use to the learning and reconstruction process.
To generate synthetic data points using autoencoders, a random latent vector in the latent space of the autoencoder is sampled and then reconstructed using the decoder network. However, since the training of autoencoders does not enforce continuity in the latent space, the decoder may not be able to reconstruct the latent vector into a real-world data point. To address this issue, variational autoencoders \cite{VAE} were introduced.

VAEs encode a data point into a latent distribution instead of a latent vector, from which a sample is drawn and used by the decoder network to reconstruct the input image. In addition to the reconstruction loss, VAEs incorporate a constraint loss on the latent space that ensures its continuity by minimizing the KL divergence of each distribution inferred from a data point to a reference distribution (usually the standard Gaussian distribution). $\beta$-VAEs \cite{beta-vae} improve on this idea by introducing a parameter that places more emphasis on the KL divergence term of the loss. However, balancing the KL divergence and the reconstruction loss can be difficult, as these terms can be contradictory and placing too much emphasis on the KL divergence can negatively impact the reconstruction process. Furthermore, even when visualizing the latent spaces of VAEs and $\beta$-VAEs that achieve good reconstruction performance, clusters can still be observed for each semantic class, indicating that the continuity of the latent space is still not guaranteed.

Wasserstein autoencoders (WAEs), introduced in \cite{wae}, aim to solve the continuity problem and introduce structure in the latent space by constraining the latent encodings to be inside a standard Gaussian distribution. This is achieved by minimizing a penalized form of the Wasserstein distance between the model and the target standard Gaussian distribution. This results in a regularizer that is different from those used by VAEs and leads to a more structured latent space. However, this constraint is only satisfied if the Wasserstein distance is zero.

Hyper-spherical Variational Autoencoders ($\mathcal{S}-\mbox{VAE}$) were introduced in \cite{SVAE} and represent an attempt to move away from euclidean latent spaces and introduce built-in topological structures as constraints on the latent space. The use of topological structures can help reduce the probability of sampling a latent vector that the decoder cannot reconstruct into a real-world sample.

One solution to address the continuity of the latent space is to fit a mixture model to the latent space of the training or validation dataset. New samples can then be drawn from the mixture model, which can lead to good performance in terms of synthetic data generation because the drawn samples are in areas where the decoder knows how to reconstruct them. However, fitting a mixture model requires the number of components to be known a priori or determined using cross-validation, which can be computationally expensive. Typically, the number of components is set to be equal to the number of classes in the dataset. However, this heuristic approach raises questions such as: What if the number of classes is unknown? What if a single class is represented by multiple clusters in the latent space? These questions remain relevant whether the latent space is euclidean or hyper-spherical, as in the case of $\mathcal{S}-\mbox{VAE}$.

In this work, we introduce the Simplex Autoencoder (Simplex AE), an autoencoder that uses a simplex topological structure on the latent space. This structure allows for a more intuitive explanation of the latent vectors, as well as a better heuristic for the number of components to use for the mixture model. In the Simplex AE, each vector in the latent space can be interpreted as the probability of having a latent feature\footnote{It is important to note that the number of features is not necessarily the same as the number of classes.}. As a result, data points with the same features will be clustered around certain vertices of the simplex, and the number of components can be set to be equal to the number of vertices in the simplex. Additionally, we propose a latent space sampling strategy based on probability mass functions that improves on the performance of mixture models. Moreover, with Simplex AE, Riemannian metric estimation on the latent space can also be considered, as in previous works such as \cite{lebanon-simplex-metric} and \cite{marco-simplex-metric}.

We summarize the novel contributions of this paper as follows :
\vspace{-.2cm}
\begin{itemize}
    \itemsep-.05cm 
    \item We propose the Simplex Autoencoder, an autoencoder with a built-in simplex topology on the latent space. This allows for an intuitively explainable latent space, a better heuristic for the number of components in a mixture model, and opens the door for Riemannian metric estimation on the latent space.
    \item We propose a latent space sampling strategy based on probability mass functions.
    \item We illustrate the properties of the Simplex AE and the proposed sampling strategy on a synthetic dataset.
    \item We validate the Simplex AE's performance on synthetic image generation on three datasets: MNIST \cite{mnist}, CIFAR-10 \cite{cifar10}, and Celeba \cite{celeba}. We evaluate the performance using classification and FID metrics.
\end{itemize}
\vspace{-.2cm}

The structure of this paper is as follows. In the first section, we review relevant concepts and introduce necessary notation. In the next section, we present the Simplex AE method, including its learning and sampling phases. In the following section, we conduct empirical evaluation of Simplex AE and compare it to other autoencoder-based approaches in the literature. Finally, we conclude the paper with a discussion of our findings and directions for future research.
\section{Background}
In this section, we provide necessary background definitions and introduce notations for the purpose of completeness.

\subsection{Autoencoders}
An \textsl{Autoencoder} ($\mbox{AE}=\{ E_\phi,D_\theta\}$) \cite{Goodfellow-et-al-2016} is a neural network architecture that consists of two algorithms: an \textsl{encoder} and a  \textsl{decoder}. The encoder, denoted by $E_\phi$, maps an input data $x$ from a space $\mathcal{X}$ to a \textsl{latent representation} $z$ in a space $\mathcal{Z}$, where $\ddim(\mathcal{Z}) \ll \ddim(\mathcal{X})$. The decoder, denoted by $D_\theta$, reconstructs the input data $\hat{x}$ using the latent representation $z$. Both the encoder and decoder have parameters $\phi$ and $\theta$, respectively.

$$z= E_\phi(x)\mbox{~~and~~}\hat{x} = D_\theta(z)\mbox{~~i.e.~~}\hat{x} = D_\theta(E_\phi(x))$$

The weights $\phi$ and $\theta$ are optimized using gradient descent to minimize the reconstruction loss, which is typically defined as the squared $\ell_2$ norm of the difference between the input data $x$ and the reconstructed data $\hat{x}$. This can be written as $\vert\vert x - \hat{x}\vert\vert_2^{2}$.

\subsection{Simplicia} 
An \textsl{$n$-simplex} is the set of vectors $z \in \mathbb{R}^{n+1}$ defined as :
$$
\mathbb{P}_n = \{ z \in \mathbb{R}^{n+1} \; |\; \forall i , z_i \geq 0 \text{ and } \vert\vert z\vert\vert_1=1\}
$$
\noindent where each $z_i$ can be seen as the probability of belonging to the class $i$ or having a feature $i$.

\subsection{Softmax function}
The Softmax function, denoted by $\sigma(\cdot)$, is a smooth approximation of the one-hot encoding of the max of a vector \footnote{A one-hot encoding of the max of a vector $v$ is a vector $u$ that indicates the position of the maximum value in $v$. $u$ is of the same dimension and contains only zeros, except at the position of the max it takes the value of one. For example, if $v=[1,2,3,9]$ then $u=[0,0,0,1]$.}. This has the benefit of being differentiable, which allows it to be used as an activation function in neural networks. Given an input vector $t \in \mathbb{R}^n$, the Softmax function outputs a probability vector $z \in \mathbb{P}_{n-1}$ on an $(n-1)$-simplex. It is defined by the following formula:
\begin{align*}
    \sigma(t) : \mathbb{R}^n & \rightarrow \mathbb{P}_{n-1}\\
    t & \rightarrow z_i = \sigma(t)_i = \frac{\exp(t_i)}{\sum_{j=1}^n \exp(t_j)}
\end{align*}

\subsection{Dirichlet distribution}
The \textsl{Dirichlet distribution} $\dir_\alpha(\cdot)$ with parameter $\alpha \in \mathbb{R}^{n+1}_+$ is a multivariate continuous distribution defined over the $n$-simplex by the following Probability Density Function (PDF):
$$
\dir_\alpha(x) = \frac{1}{B(\alpha)} \prod_{i=1}^{n+1}x_i^{\alpha_i-1}; \; x\in \mathbb{R}^{n+1}
$$
\noindent where $B(\alpha)$ is the multivariate $\beta$ function serving as a normalizing constant. We recall that $B$ is the following special function defined for $z\in\mathbb{C}^n$ such that $\forall i \in \{1, \dots, n\},  \Re(z_i)>0$:

$$
B(\alpha) = \frac{\prod_{i=1}^{n+1} \Gamma(\alpha_i)}{\Gamma\left( \sum_{i=1}^{n+1} \alpha_i \right)}
\mbox{~~where~~}\Gamma(z) = \int_0^\infty t^{z-1}e^{-t}\,dt$$

\subsection{Logistic normal distribution}
The \textsl{Logistic normal distribution} $\log\mathcal{N}(x;\mu,\Sigma)$ is a probability distribution defined on the $n$-simplex $\mathbb{P}_n$ with the following PDF :

$$
\log\mathcal{N}(x;\mu,\Sigma) = \frac{1}{\left\vert (2\pi)^{n}\Sigma\right\vert^{\frac{1}{2}}} \frac{1}{\prod^n_{i=1} x_i} \exp \left[ -\frac{1}{2} \left(\log\left(\frac{x_{-n}}{x_n}\right) - \mu\right)^\intercal \Sigma^{-1} \left(\log\left(\frac{x_{-n}}{x_n}\right) - \mu\right)\right]
$$

\noindent $x_{-n}\in \mathbb{R}^n$ denotes the vector $x \in \mathbb{R}^{n+1}$ with the last component removed. $\mu \in \mathbb{R}^{n}$ and $\Sigma \in \mathbb{R}^{(n, n)}$ are the mean and covariance parameters, respectively, of this distribution.

To sample a vector $x \in \mathbb{R}^{n+1}$ from the logistic normal distribution, we first sample $y\in \mathbb{R}^n$ from a multivariate normal distribution $\mathcal{N}(y;\mu,\Sigma)$. Then, we apply the following logistic transformation: 
\begin{equation}
    x = \left[\frac{\exp(y_1)}{1+\sum_{i=1}^n \exp(y_i)}, \dots, \frac{\exp(y_n)}{1+\sum_{i=1}^n \exp(y_i)},\frac{1}{1+\sum_{i=1}^n \exp(y_i)} \right]^\intercal.
    \label{projection to simplex}
\end{equation}

The inverse of the logistic transformation allows us to transform a sample from the logistic normal distribution to a sample from a Gaussian distribution. It is given by:

\begin{equation}
    y = \left[ \log\left(\frac{x_1}{x_{n+1}}\right), \dots, \log\left(\frac{x_n}{x_{n+1}}\right) \right]^\intercal.
    \label{projection to euclidean}
\end{equation}

\subsection{Mixture models}
\label{mixture models}
\textsl{Mixture models} assume that data may contain sub-populations. Therefore, if we model each sub-population $i$ by a probability distribution $p_i(\cdot\vert\psi_i)$ with parameters $\psi_i$, the probability distribution of the entire dataset can be expressed as a sum of the probability distributions $p_i(\cdot\vert\psi_i)$, weighted by weights $\alpha_i$ where $\sum_i \alpha_i =1$. In this way, the probability of any data point in the dataset can be written as~: 
$$
p(x) = \sum_i \alpha_i p_i(x\vert\psi_i).
$$
The base probability distributions $p_i(\cdot\vert\psi_i)$ can be chosen to be Gaussian distributions, in which case the parameter set $\psi_i$ includes the mean and variance of the distribution $p_i(\cdot\vert\psi_i)$. In this case, the distribution $p(x)$ would represent a \textsl{Gaussian mixture model.}

A logistic normal mixture model on an $n$-simplex can be defined by setting the $p_i(\cdot\vert\psi_i)$ to be logistic normal distributions. This model is analogous to a Gaussian mixture model. The parameters of the logistic normal mixture model can be estimated by projecting the data from the simplex to Euclidean space using Equation \ref{projection to euclidean} and fitting a Gaussian mixture model to the projection using the expectation-maximization (EM) algorithm\footnote{An expectation-maximization (EM) algorithm is a statistical method for finding the parameters of a statistical model that involve unobserved variables. The EM algorithm iterates two steps : an expectation (E) step and a maximization (M) step. During the E step, the algorithm calculates the expected value of the log-likelihood function using the current estimates of the parameters. In the M step, the algorithm updates the parameters by finding the values that maximize the expected log-likelihood calculated in the E step \cite{EM}.}. To sample from the logistic normal mixture model, one can sample from the fitted Gaussian mixture model and then project the data back onto the simplex using Equation \ref{projection to simplex}.

\subsection{Wasserstein distance}
\textsl{Optimal transport} involves comparing and measuring the distance between probability distributions. The \textsl{Wasserstein distance} is a widely used metric for this purpose, and it can be calculated between two distributions, $p$ and $q$, as follows:
$$
W_\xi(p,q) = \left( \inf_{\gamma \in \mathcal{P}(p(x), q(x^\prime))} \mathbb{E}_{\gamma(x, x^\prime)} [d^\xi(x,x^\prime)]  \right)^\frac{1}{\xi}.
$$

Since the computation of the infimum\footnote{The infimum of a subset $S$ of a partially ordered set $P$ is the greatest $p\in P$ that is less than or equal to each element of $S$, if such an element exists.} is in most cases computationally infeasible, entropy regularized optimal transport was introduced in \cite{sinkhorn_distances} along with the Sinkhorn algorithm and further improved in \cite{genevay:entropy-reg-ot} which allowed the approximation of the Wasserstein distance in reasonable time.
\section{Our Contribution}
In this section, we introduce the concept of Simplex Autoencoders and propose various sampling algorithms for generating synthetic data using Simplex AEs, including strategies based on mixture models and probability mass functions.

\subsection{Simplex Autoencoders}
\label{simplex ae}
Simplex Autoencoders (Simplex AEs) incorporate an $n$-simplex topology into the latent space of an autoencoder with no additional training parameters, where $n$ is the dimension of the latent space. This is achieved by applying the Softmax function $\sigma(\cdot)$ to the output of the encoder network $E_\phi$. The operations applied to the input data $x$ in a Simplex AE can be summarized by the following equations:
\begin{align}
    t       &= E_\phi(x)\\ \nonumber
    z       &= \sigma(t)\\ \nonumber
    \hat{x} &= D_\theta(z)=D_\theta(\sigma(E_\phi(x)))
\end{align}

Modeling the latent space as a simplex rather than a hypercube results in latent vectors that represent probabilities and live on a simplex. Each element of the latent vector $z$ represents the probability of having a specific latent feature, and the sum of all elements of the latent vector is equal to one. This choice is motivated by the fact that an $n$-hypercube is isomorphic to an $(n-1)$-simplex, and we want to reduce the volume of the latent topological structure, which will directly reduce the number of vectors that cannot be reconstructed by the decoder $D_\theta$.

{Data with similar features will have latent probability vectors that are close to each other and concentrated near the vertices of the $n$-simplex. Therefore, if we sample from the latent space using a mixture model, a natural choice for the number of components is the dimensionality of the latent space, $n$. This is illustrated in the simulated setting in Section \ref{results_synthetic}. In addition, the compactness of the latent space means that the decoder will learn to reconstruct most data points in the latent space, and therefore most synthetic data points will follow the distribution of the training data, provided the most optimal sampling strategy is used (see Section \ref{results_synthetic} and \ref{results_benchmark}).

Furthermore, previous works such as \cite{lebanon-simplex-metric} and \cite{marco-simplex-metric} define ways to learn the Riemannian metric on a simplex and perform density estimation. These methods could also potentially be used in conjunction with Simplex AE, but this possibility is not explored in the present article and is reserved for future work.

\subsection{Learning phase}
To learn the weight parameters $\phi$ and $\theta$, the Simplex AE is trained to minimize the reconstruction loss, which is modeled by the squared $\ell_2$ norm of the difference between $x$ and $\hat{x}$. In addition, to impose further structure and more control over the latent space, we minimize the 2-Wasserstein distance on the latent space to a reference distribution on the simplex, chosen to be the Dirichlet distribution $\dir_\alpha(\cdot)$. The Dirichlet distribution is chosen over other distributions, such as the logistic normal distribution, because it has fewer hyperparameters. A hyperparameter $\lambda$ is used to balance the reconstruction loss and the 2-Wasserstein distance so that they are on the same order. The final training loss of the Simplex AE is:
\begin{equation}
    \mathcal{L}(x, \hat{x}) = \vert\vert x-\hat{x} \vert\vert_2^{2} + \lambda W_2(E_\phi(x), \dir_\alpha(\cdot)).
\end{equation}

To approximate the 2-Wasserstein distance, we use Sinkhorn's algorithm, as described in the works of Genevay et al. \cite{genevay:entropy-reg-ot} and Feydy et al. \cite{geomloss}, by comparing samples drawn from the Dirichlet distribution to the latent vectors $z$ of the input data $x$.

\subsection{Sampling phase}
\label{sampling methods}
To generate synthetic data, we propose several approaches for sampling a vector $z$ from the latent space of the Simplex AE and using the decoder to reconstruct it into a data point. These approaches include uniform sampling, $\alpha$-sampling, logistic Gaussian mixture sampling, and probability mass sampling. We will compare and evaluate these sampling approaches in the following section.

\subsubsection{Uniform sampling}
This sampling strategy involves randomly sampling a vector $z$ from $\dir_1(\cdot)$, with the parameter $\alpha$ in the $\dir_\alpha(\cdot)$ distribution set to a vector of ones.

\subsubsection{\texorpdfstring{$\alpha$}{}-sampling}
In this strategy, we sample from the Dirichlet distribution using the parameters that were used to calculate the Wasserstein distance during the training phase. $\alpha$-sampling should produce samples that are more representative of the training data, since during the learning phase we minimized the distance between the latent encodings and this distribution. Therefore, the latent space should be distributed according to this distribution.

\subsubsection{Logistic normal mixture sampling}
This sampling strategy is able to identify sub-populations in the latent space and model them accurately. Since the Simplex AE approach provides a good approximation of the number of components, we will test a number of components equal to the number of classes in the dataset (as is standard in the literature) and a number of components equal to the number of simplex vertices, i.e. the number of latent space dimensions. To use this strategy, we follow the steps outlined in Section \ref{mixture models} to fit the Logistic normal mixture model to the latent space of the validation data.

\subsubsection{Probability mass function sampling}
Estimating the probability mass allows us to identify regions of the latent space where the embeddings of the training data are located. Consequently, sampling from these regions increases the probability that the synthetic data points will follow the same distribution as the training data. This strategy is more feasible on a latent space in the form of a simplex, since it is a compact topological space with boundaries, than on a Euclidean latent space. This strategy is implemented by uniformly partitioning the range $[0,1]$ along each dimension into $k$ bins, resulting in a total of $k^n$ bins for the whole latent space. Then, a weighted sampling of a bin is performed, where the weights are the number of data points in each bin. Finally, to sample a vector $z$, we uniformly sample from the selected bin.

Probability mass function (PMF) sampling relies on space partitioning and can result in an exponentially increasing number of bins as the latent dimension $n$ increases, due to the $k^n$ partitions created by the adopted partitioning method. To avoid this problem and maintain performance, we only preserve bins that contain data points. This is equivalent to keeping all bins, as bins with no data samples will never be selected.


\section{Experiments}
This section presents experimental results that validate the proposed Simplex AE method (described in Section \ref{simplex ae}) and compare the different sampling strategies introduced in Section \ref{sampling methods}.

\subsection{Experimental setup}
To train the models for this work, we used the Pytorch \cite{pytorch} and Pytorch Lightning \cite{pytorch_lightning} libraries in conjunction with Torchvision \\ \cite{torchvision} to obtain the training benchmark datasets and pretrained models. The training was done on an Intel Xeon Silver 4214R CPU with 8 cores and 16GB of RAM. It takes about 4 minutes to train a Simplex AE on the MNIST \cite{mnist} and CIFAR-10 \cite{cifar10} datasets for one epoch, and 15 minutes to calculate the FID. We conducted experiments on the Celeba dataset using Nvidia Tesla V100 GPUs with 32 GB of VRAM, which are available in the MesoPSL computing cluster. With these GPUs, training on the Celeba dataset took approximately 6 minutes per epoch, while computing the FID required approximately 30 minutes.

The Sklearn library \cite{scikit-learn} was used not only to generate the synthetic dataset for this work, but also to implement the different sampling strategies in conjunction with the NumPy library \cite{numpy}. The Matplotlib library \cite{matplotlib} was used to generate the figures in this work.

\subsection{The synthetic dataset}
\label{results_synthetic}
To illustrate our approach, we use a controlled synthetic dataset containing $20$ features, of which three are informative and a varying number of classes $(3,7,8)$. The dataset was constructed using the \texttt{make\_classification} functionality of the sklearn library \cite{scikit-learn}. We use a spacing of five between classes and one cluster per class. The dataset contains $20,000$ samples, with $10,000$ for training, $5,000$ for validation, and $5,000$ for testing. Figure \ref{fig:synthetic_dataset_latent_space} shows the latent encodings of the test set of a Simplex AE trained for twenty epochs with $\lambda=100$ and an architecture described in Appendix \ref{sythetic_dataset_ae_archi}.

\begin{figure}[ht]
    \centering
     \begin{subfigure}{0.32\textwidth}
         \centering
         \includegraphics[width=\textwidth]{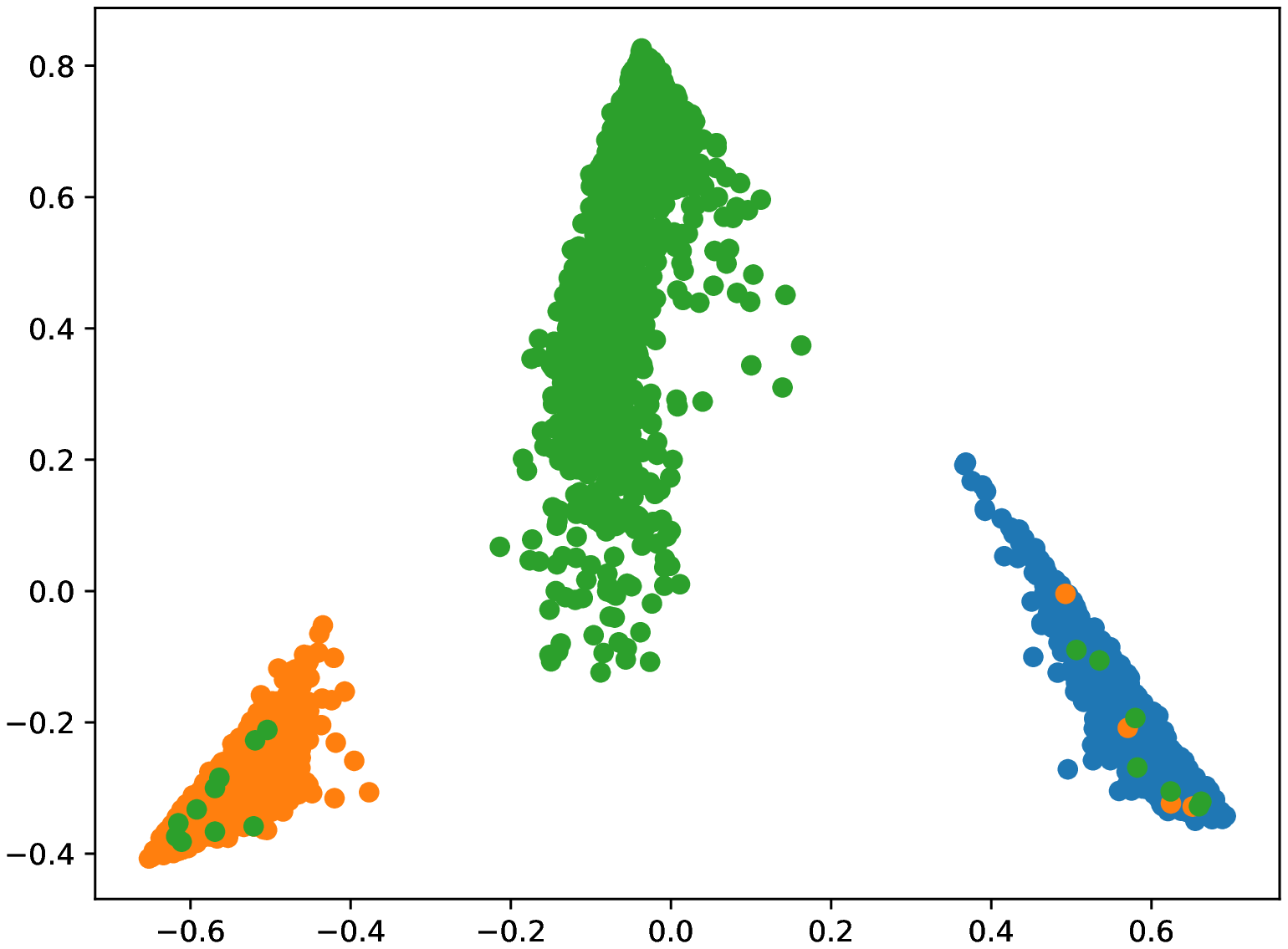}
         \caption{3 classes, $\alpha=0.3$.}
     \end{subfigure}
     \hfill
     \begin{subfigure}{0.32\textwidth}
         \centering
         \includegraphics[width=\textwidth]{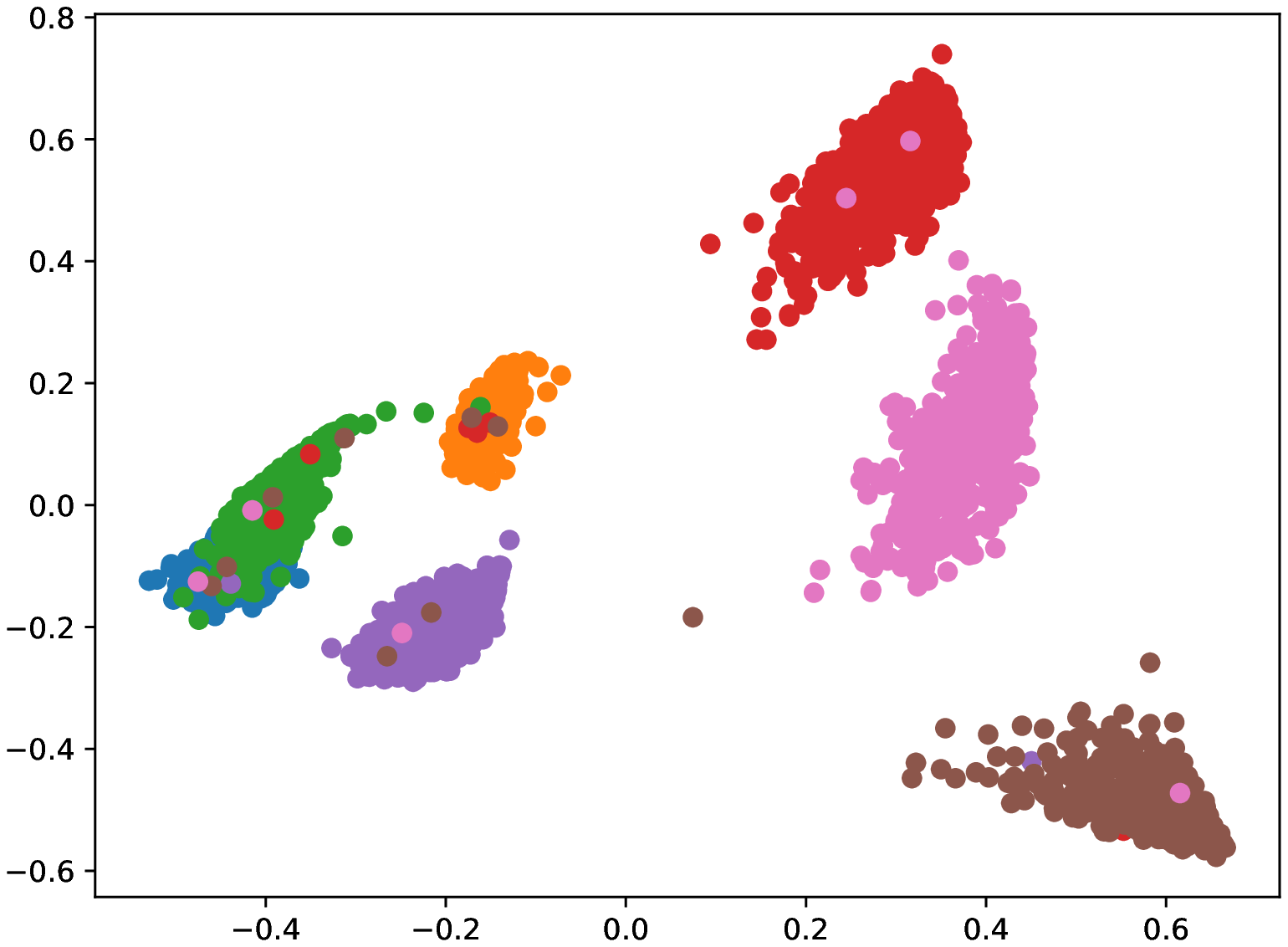}
         \caption{7 classes, $\alpha=0.3$.}
     \end{subfigure}
     \hfill
     \begin{subfigure}{0.32\textwidth}
         \centering
         \includegraphics[width=\textwidth]{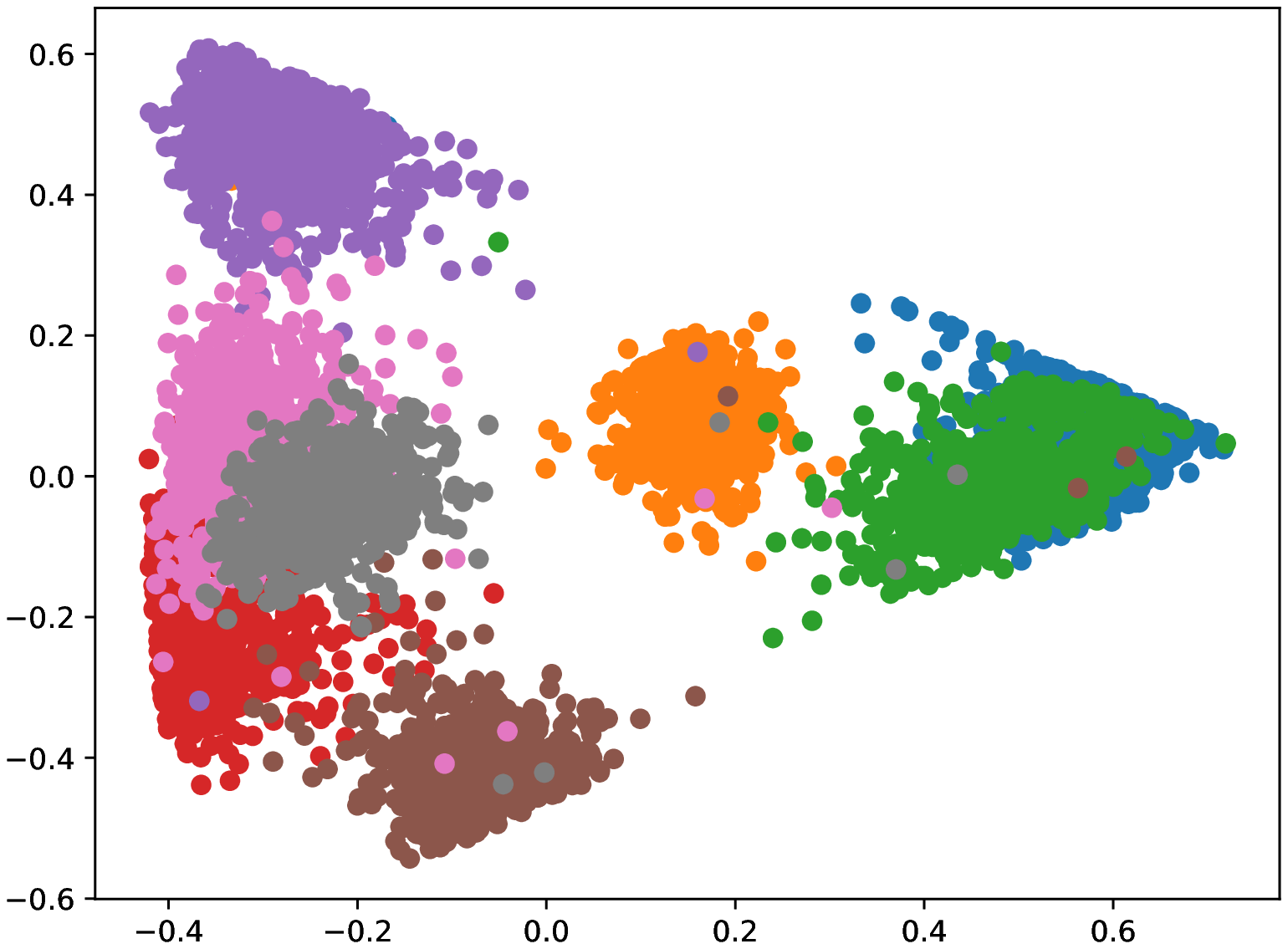}
         \caption{8 classes, $\alpha=0.3$.}
     \end{subfigure}
     \begin{subfigure}{0.32\textwidth}
         \centering
         \includegraphics[width=\textwidth]{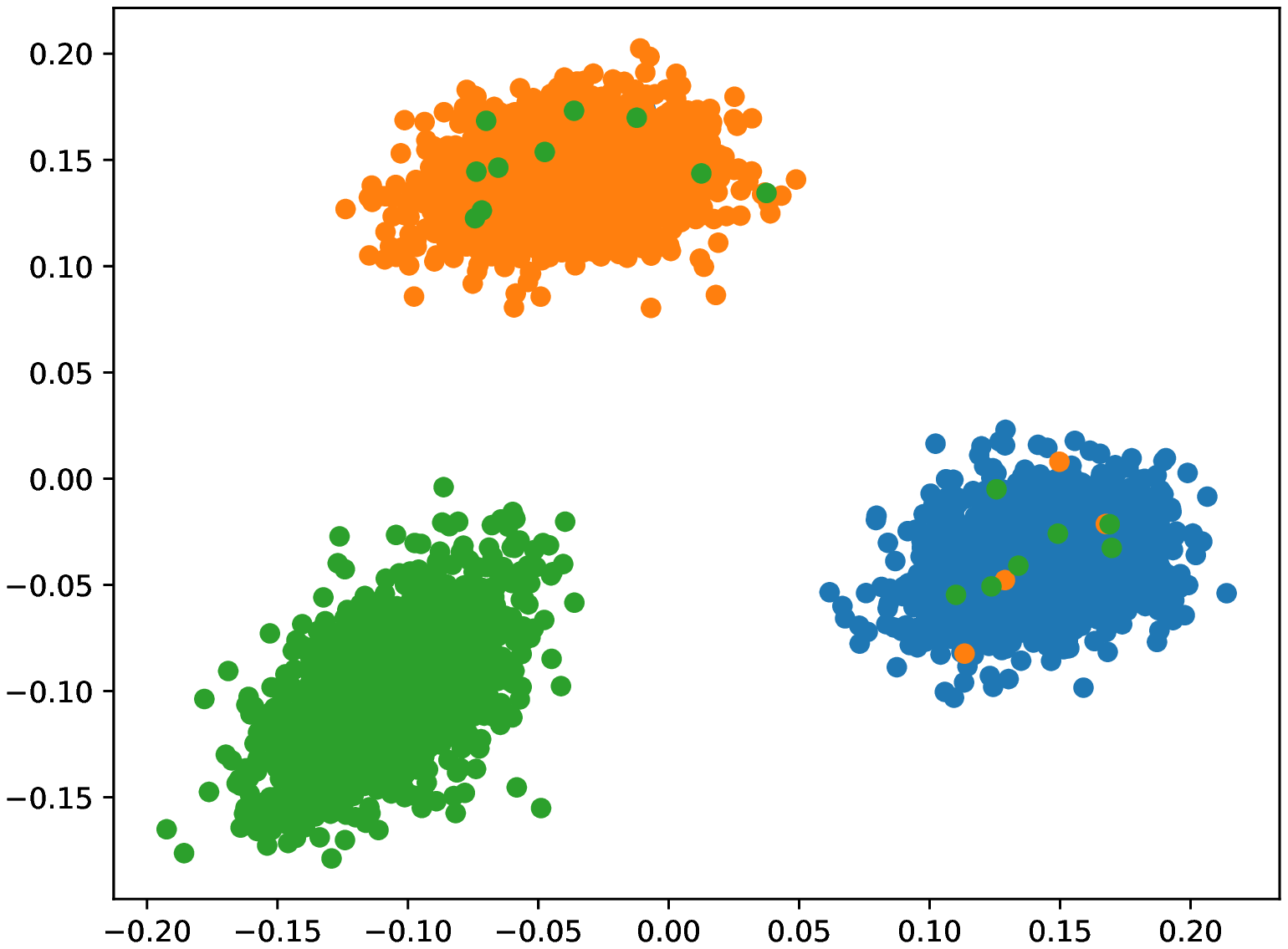}
         \caption{3 classes, $\alpha=30$.}
     \end{subfigure}
     \hfill
     \begin{subfigure}{0.32\textwidth}
         \centering
         \includegraphics[width=\textwidth]{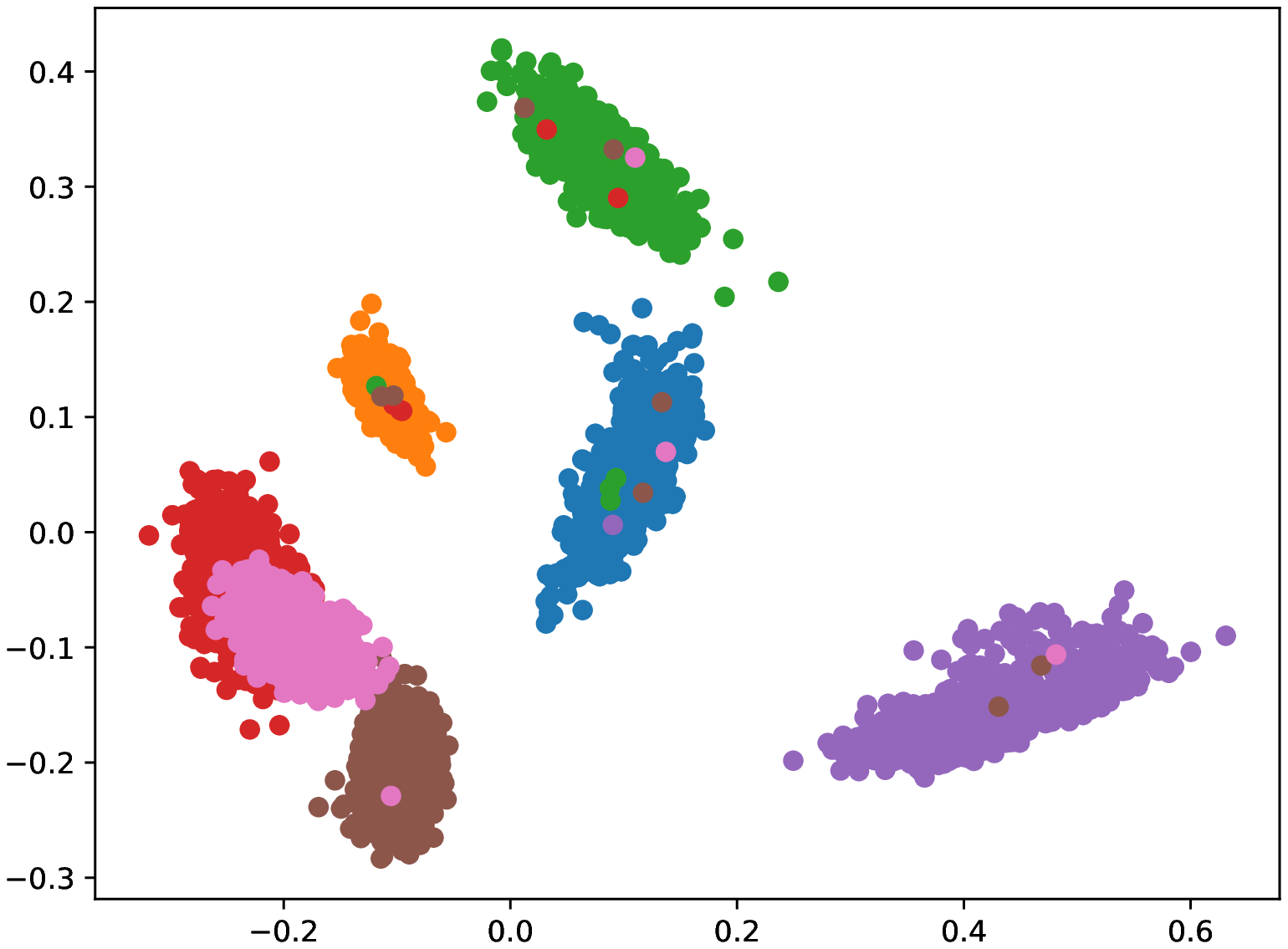}
         \caption{7 classes, $\alpha=30$.}
     \end{subfigure}
     \hfill
     \begin{subfigure}{0.32\textwidth}
         \centering
         \includegraphics[width=\textwidth]{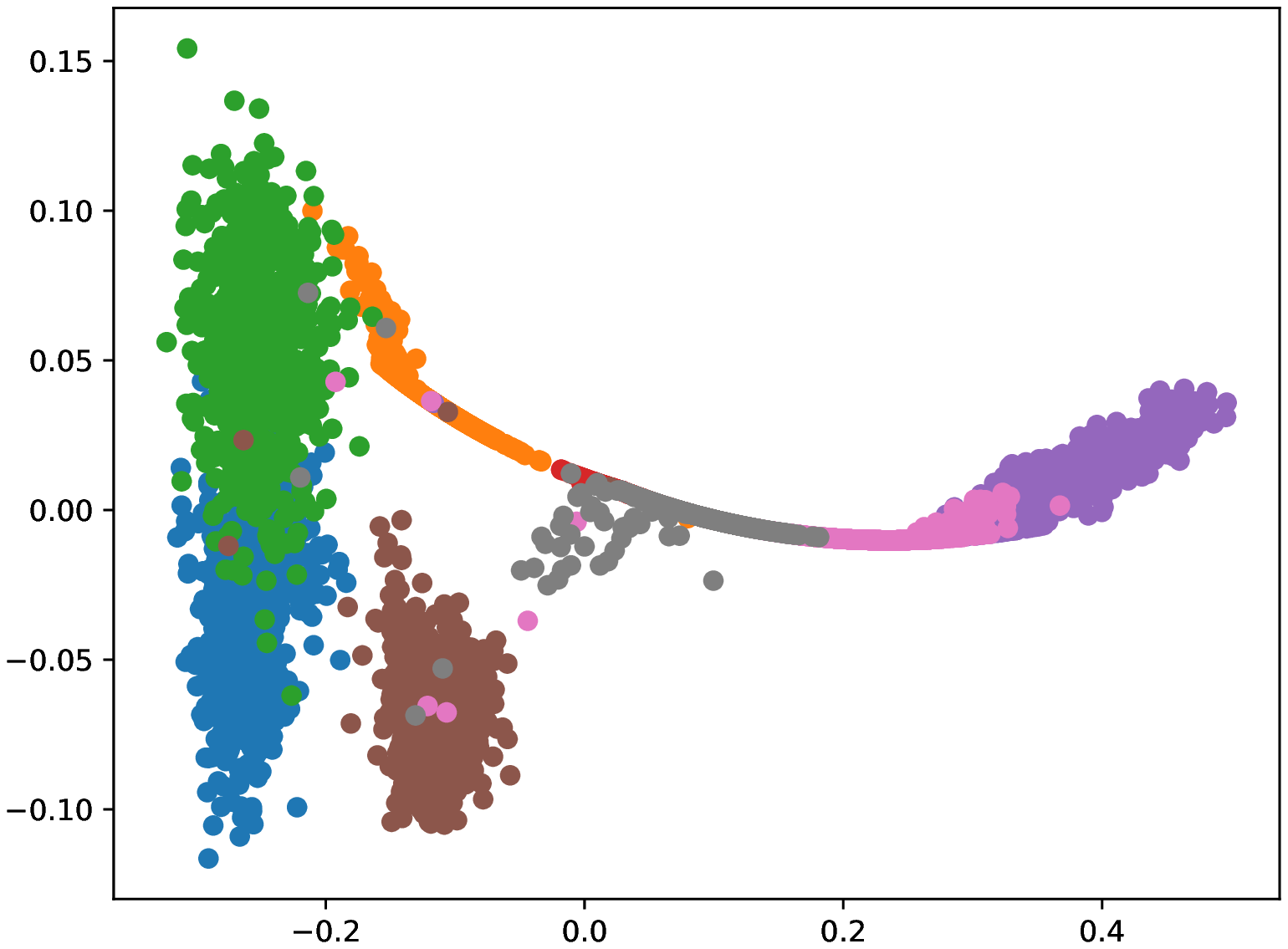}
         \caption{8 classes, $\alpha=30$.}
     \end{subfigure}
     
    \caption{Latent space of Simplex AE on synthetic dataset. The indicated $\alpha$ values represents a vector in $\mathbb{R}^{3}$ having that particular value. Each color represents a class in the dataset. Figure better
viewed in color.}
    \label{fig:synthetic_dataset_latent_space}
\end{figure}

Figure \ref{fig:synthetic_dataset_latent_space} demonstrates not only the effect of the parameter $\alpha$ on the latent space distribution of the Dirichlet distribution, but also the number of classes that can be represented in a simplex. As $\alpha$ decreases, the latent space becomes more concentrated near the vertices of the simplex, whereas higher values result in a more centralized distribution. Up to seven classes, distinct clusters are visible in the 3-simplex. However, when eight classes are used, the structure becomes obscured. This is also evident in the loss value, which nearly doubles when the number of classes increases from seven to eight (see Figure \ref{fig:bar_plot_loss}). Experiments with a 4-simplex showed that up to 15 classes can be packed into the space. This suggests that the maximum number of clusters that can be encoded in an $n$-simplex is $2^{n+1}-1$, but this may vary depending on the spacing between the clusters. Further investigation of this topic is left for future work.

\begin{figure}[H]
    \centering
    \includegraphics[width=.5\textwidth]{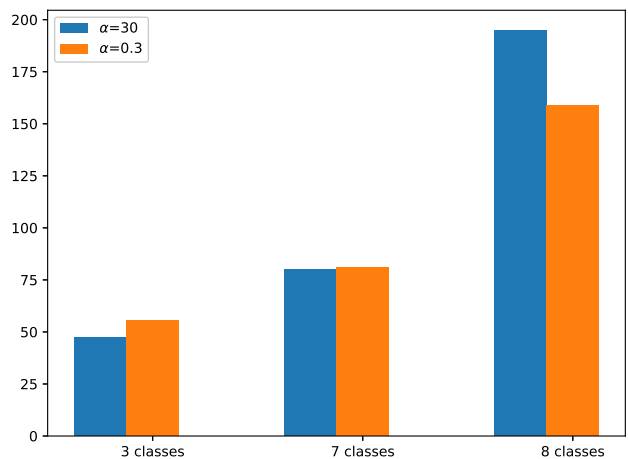}
    \caption{Bar chart showing the loss function value of the Simplex AE on a synthetic dataset when using a $\mathbb{P}_2$ latent space and varying the number of classes in the dataset. Figure better
viewed in color.}
    \label{fig:bar_plot_loss}
\end{figure}

\begin{figure}[H]
    \centering
     \begin{subfigure}{0.24\textwidth}
         \centering
         \includegraphics[width=\textwidth]{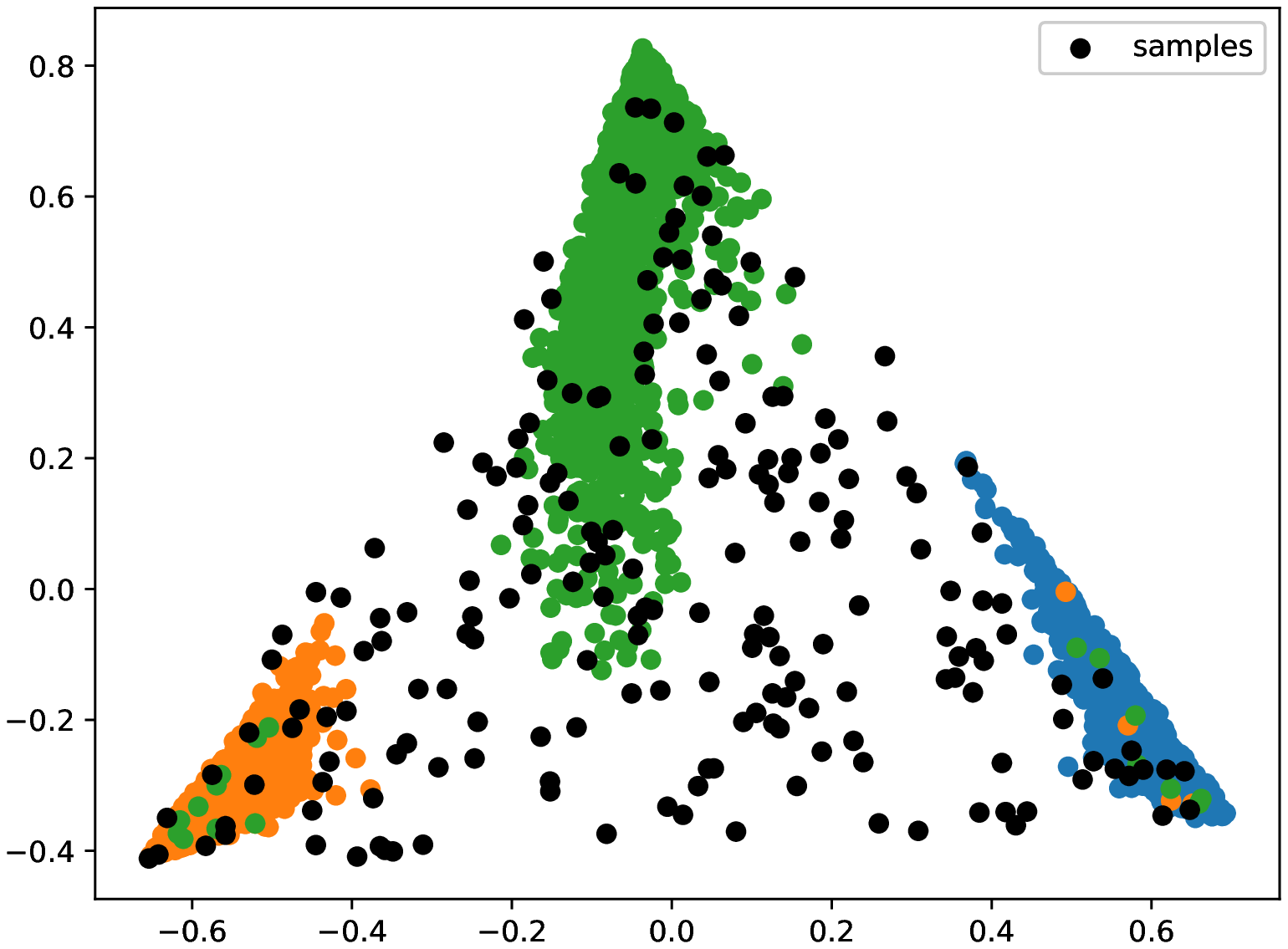}
         \caption{Uniform sampling.}
     \end{subfigure}
     \hfill
     \begin{subfigure}{0.24\textwidth}
         \centering
         \includegraphics[width=\textwidth]{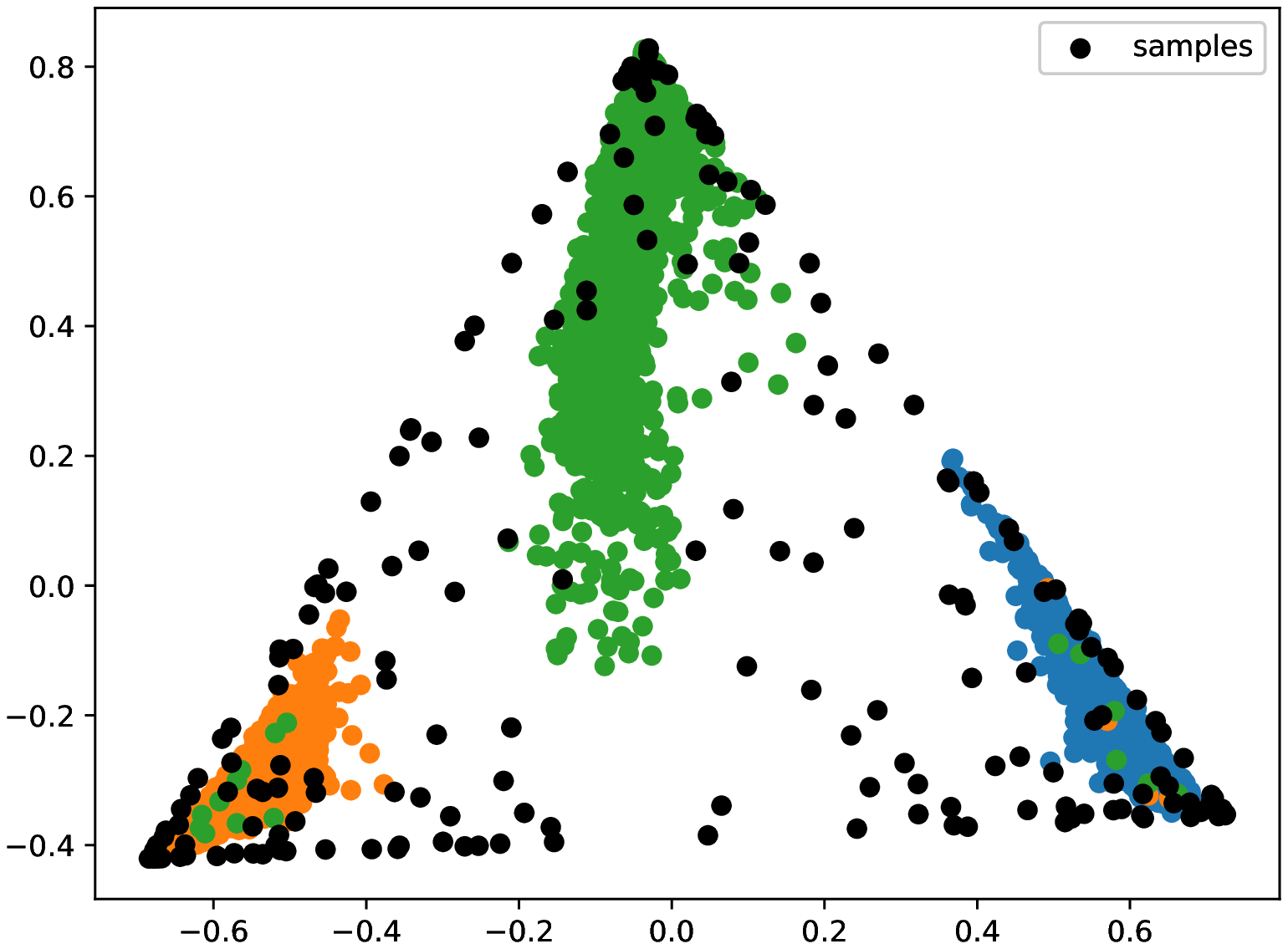}
         \caption{$0.3$-sampling.}
     \end{subfigure}
     \begin{subfigure}{0.24\textwidth}
         \centering
         \includegraphics[width=\textwidth]{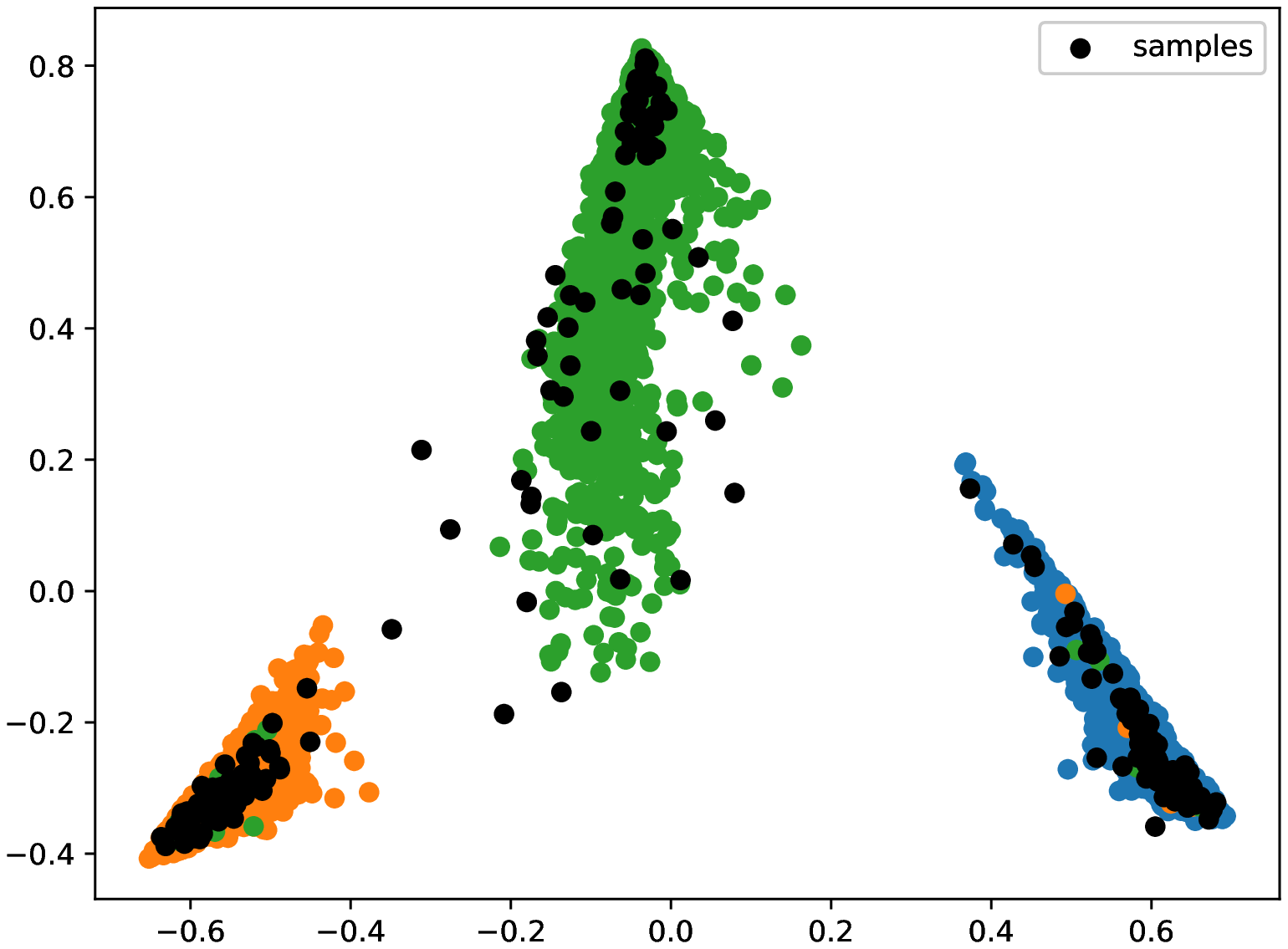}
         \caption{MM sampling.}
     \end{subfigure}
     \hfill
     \begin{subfigure}{0.24\textwidth}
         \centering
         \includegraphics[width=\textwidth]{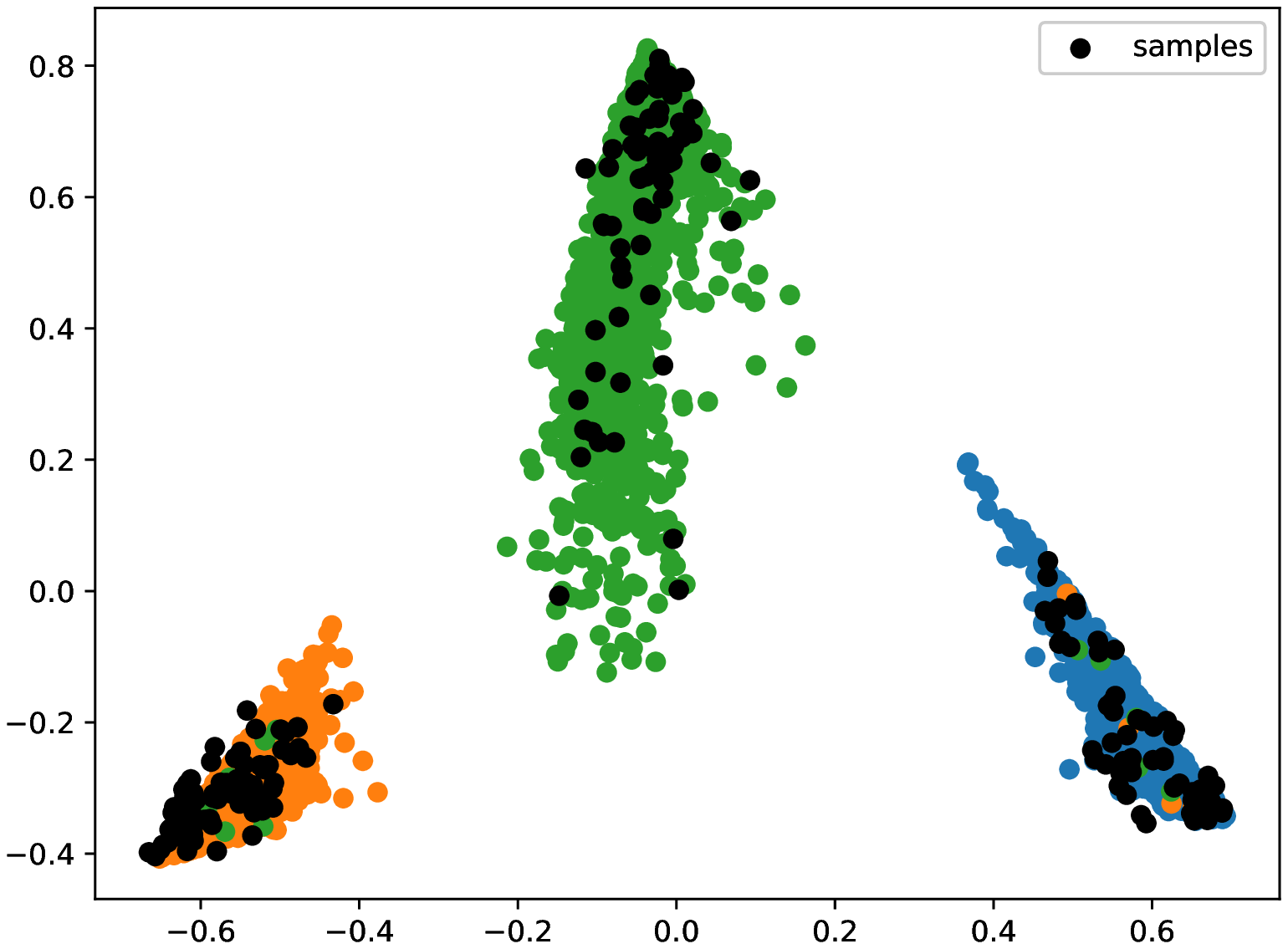}
         \caption{PMF sampling.}
     \end{subfigure}
    \caption{Comparison of sampling strategies. MM sampling refers to Logistic normal mixture sampling with three components. PMF sampling refers to probability mass function sampling with $k=20$. Figure better viewed in color.}
    \label{fig:sampling_synthetic}
\end{figure}

Figure \ref{fig:sampling_synthetic} shows sampled latent vectors using the sampling methods introduced earlier. For this figure, we used a Simplex AE trained on three classes with $\alpha$ equal to $0.3$. We can see that both the logistic normal mixture and probability mass function sampling produce samples that are within the clusters of the latent space. This means that the reconstructed data will be similar to the training data. In contrast, the uniform and alpha sampling produce many points outside the clusters, which the decoder will not be able to reconstruct accurately. In the next section, where we evaluate the model on benchmark datasets, we will only use the logistic normal mixture and probability mass function sampling strategies.

\subsection{Benchmark datasets}
\label{results_benchmark}
\subsubsection{Training parameters}

We evaluated the performance of our algorithm using the MNIST, CIFAR-10, and Celeba datasets, in terms of the following metrics:

\begin{itemize}
    \item \textsl{Fréchet Inception Distance} (FID) \cite{FID}, a widely used metric for evaluating the quality of synthetic images generated by generative models such as GANs and AEs by comparing the distribution of generated images to that of a set of real-world images, with a score of 0 indicating that the distribution of synthetic images match that of real ones.
    
    \item \textsl{$K$-Nearest Neighbors} (KNN) Classification Accuracy \cite{knn1, knn2} to assess the quality of the produced embeddings for downstream tasks such as classification, by training the model on the latent embeddings of images from the validation set and scoring it on the images of the test set of each dataset.
    
    \item \textsl{Peak Signal-to-Noise Ratio} (PSNR), to evaluate the quality of the reconstruction process and quantify the amount of information lost, which is inversely correlated to the $\ell_2$ norm, with a higher score indicating better reconstruction quality, calculated on the test images of each dataset.
    
\end{itemize}

The datasets were pre-processed to have three channels and a resolution of $64\times 64$ except for the Celebra dataset where we keep the original resolution. The Simplex AE was trained for $100$ epochs using the Adam optimizer with a learning rate of $10^{-4}$, a $\lambda$ value of $100$ for the training loss, a neural architecture described in the Appendix \ref{benchmark_dataset_ae_archi} and a batch size of $64$. For the MNIST dataset, which is a simpler dataset, we train our model for only 50 epochs with a learning rate of $10^{-3}$. The other hyperparameters are unchanged. We found that all models converged and we provide samples of synthetic images from the best-performing model for each dataset in the Appendix \ref{img_samples}. We also provide a study of the hyperparameter $\alpha$ in Appendix \ref{alpha_study}.

\subsubsection{MNIST}
Table \ref{tab:mnist_fid_results} illustrates the FID, KNN classification accuracy and PSNR scores in decibel (dB) of a Simplex AE trained on the MNIST dataset with $\alpha=30$. 

\begin{table}[H]
    \centering
    \begin{tabular}{|c|c|c|cc|c| c|}\toprule
~~\textbf{dim}~~&~~\textbf{FID - MM-10 $\downarrow$}~~&~~\textbf{FID - MM-\#dim $\downarrow$}~~&
\multicolumn{2}{|c|}{~~\textbf{FID - PMF $\downarrow$}~~} 
& ~~\textbf{KNN accuracy $\uparrow$}~~&~~\textbf{PSNR $\uparrow$}\\\midrule
        3 & 23.87 & 23.32 & 22.76 & $k=2\phantom{1}$ & 77.05\% & 15.28 \\
        4 & 14.71 & 14.57 & 14.45 & $k=2\phantom{1}$ & 87.98\% & 16.19 \\
        8 & 8.92 & 8.99 & 8.13 & $k=10$ & 94.78\% & 19.63 \\
        16 & 6.03 & 6.07 & 5.46 &$ k=10$ & 96.44\% & 24.12 \\
        32 & \textcolor{red}{5.75} & \textcolor{red}{5.21} & \textcolor{blue}{4.59} & $k=10$ & \textcolor{red}{97.20\%} & \textcolor{blue}{27.77} \\
        64  & \textcolor{blue}{5.97} & \textcolor{blue}{5.36} & \textcolor{red}{4.29} & $k=10$ & \textcolor{blue}{96.55\%} & \textcolor{red}{31.16} \\\bottomrule
    \end{tabular}
    \caption{FID, KNN classification and PSNR results on the MNIST dataset. FID-MM-10 and FID-MM-\#dim denote logistic normal mixture sampling with a number of components equal to ten and the latent space dimension respectively. FID-PMF denotes probability mass function sampling, where the number of bins $k$ is indicated for each model. \textcolor{red}{Red} values represent the best and \textcolor{blue}{blue} the second best.}
    \label{tab:mnist_fid_results}
\end{table}
Table \ref{tab:mnist_fid_results} illustrates that the Simplex AE model achieves a high accuracy in classifying images using the KNN method, indicating that the resulting latent space effectively captures important class information, even when using a relatively simple classifier such as KNN. Additionally, the FID results obtained with MM sampling with a number of components equal to the dimension of the latent space are comparable to or better than those obtained with MM sampling with a number of components equal to the number of classes (10). This result is significant, because it confirms that Simplex AE provides a reliable heuristic for determining the number of components in the mixture model. As for the FID results when PMF sampling is used, we observe that, for a reasonable latent space size, we can further improve the performance of MM sampling by up to one FID. The PSNR scores obtained are also particularly high, indicating that the input images are reconstructed to a high degree of accuracy.

Figure \ref{fig:k_study_mnist} illustrates the evolution of the FID as a function of $k$, the number of partitions in the PMF sampling. From this figure, we see that when the latent space dimension ($\dim$) is small, the FID does not improve as $k$ increases. This is likely due to the low capacity of the latent space and its inability to contain a large amount of information. However, starting from $\dim=8$, we see improvements and the FID curve starts to decrease as $k$ increases. The best FID results were obtained with $k=10$ and $\dim=64$.

\begin{figure}[H]
    \centering
    \includegraphics[width=.8\textwidth]{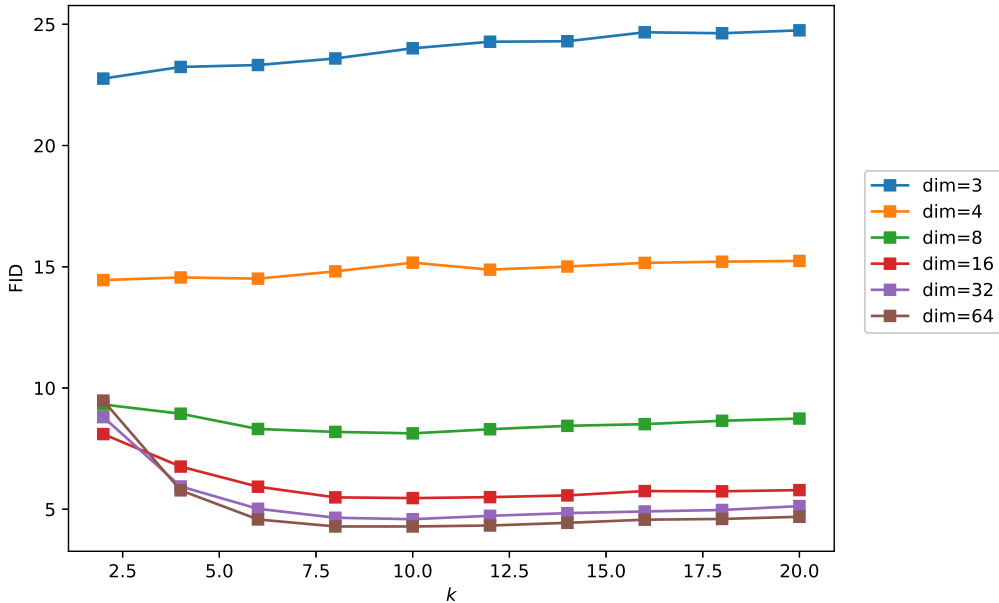}
    \caption{Evolution of the FID as a function of $k$ in PMF sampling on the MNIST dataset.}
    \label{fig:k_study_mnist}
\end{figure}

\subsubsection{CIFAR-10}
Table \ref{tab:cifar_fid_results} illustrates the FID, KNN classification accuracy and PSNR scores in decibel (dB) of a Simplex AE trained on the CIFAR-10 dataset with $\alpha=30$.

\begin{table}[H]
    \centering
    \begin{tabular}{|c|c|c|cc|c|c|}\toprule
~~\textbf{dim}~~&~~\textbf{FID - MM-10 $\downarrow$}~~&~~\textbf{FID - MM-\#dim $\downarrow$}~~&\multicolumn{2}{|c|}{~~\textbf{FID - PMF $\downarrow$}~~}& ~~\textbf{KNN accuracy $\uparrow$}~~&~~\textbf{PSNR $\uparrow$}~~\\\midrule
        3 & 30.78 & 30.93 & 28.68 & $k=2$ & 16.49\% & 14.85 \\
        4 & 25.79 & 25.80 & 22.92 & $k=2$ & 19.71\% & 15.43 \\
        8 & 17.37 & 17.45 & 15.30 & $k=2$ & 26.78\% & 16.81 \\
        16 & 15.95 & 15.94 & 14.70 & $k=4$ & \textcolor{red}{30.52\%} & 18.37 \\%
        32 & 15.61 & 15.62 & 14.43 & $k=2$ & \textcolor{blue}{29.09\%} & 19.04 \\
        64 & \textcolor{red}{15.35} &  \textcolor{blue}{15.25} & 13.93 & $k=2$ & 25.73\% & 19.57 \\
        128 & 16.74 & 16.15 & \textcolor{red}{13.55} & $k=2$ & 25.70\% & \textcolor{red}{20.93} \\
        256 &  \textcolor{blue}{15.54} & \textcolor{red}{14.90} &  \textcolor{blue}{13.62} & $k=4$ & 25.77\% & \textcolor{blue}{20.05} \\\bottomrule
    \end{tabular}
    \caption{FID, KNN classification and PSNR results on the CIFAR-10 dataset. FID-MM-10 and FID-MM-\#dim denote logistic normal mixture sampling with a number of components equal to ten and the latent space dimension respectively. FID-Mass denotes probability mass function sampling, where the number of bins $k$ is indicated for each model. \textcolor{red}{Red} values represent the best and \textcolor{blue}{blue} the second best.}
    \label{tab:cifar_fid_results}
\end{table}

Table \ref{tab:cifar_fid_results} shows that when comparing the sampling of a mixture model with ten components to a mixture model with a number of components equal to the latent space dimension, the FID values are equivalent in almost all cases on the CIFAR-10 dataset. This demonstrates that the heuristic of using a number of components equal to the number of latent dimensions provides a good approximation on the CIFAR-10 dataset. Additionally, PMF sampling allows a gain of $2$ to $3$ in FID compared to the mixture model sampling methods. The KNN classification and PSNR scores on the CIFAR-10 dataset are lower in comparison to those obtained on the MNIST dataset. This can likely be attributed to the lower resolution and increased complexity of the images in the CIFAR-10 dataset, making it a more challenging task.

Moreover, we conducted a study to investigate the effect of the hyperparameter $k$ on the PMF sampling in the Simplex AE on the CIDAR-10 dataset. The results are shown in Figure \ref{fig:k_study_cifar}. It can be seen from the figure that $k$ has minimal influence on the FID performance of the Simplex AE. Instead, the latent space dimension seems to be the main factor that determines the FID score.

\begin{figure}[H]
    \centering
    \includegraphics[width=.8\textwidth]{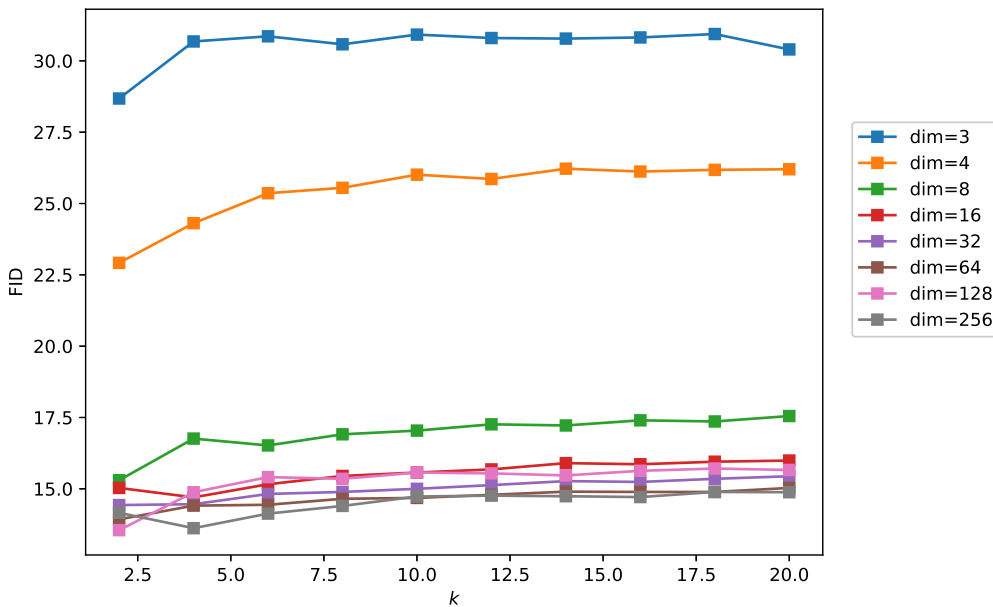}
    \caption{Evolution of the FID as a function of $k$ in PMF sampling on the CIFAR-10 dataset.}
    \label{fig:k_study_cifar}
\end{figure}

\subsubsection{Celeba}
The Simplex AE model has previously been evaluated on low-resolution image datasets. In this section, we examine its performance on a high-resolution dataset, Celeba \cite{celeba}, by using the original image sizes. Table \ref{tab:celeba_fid_results} presents the FID score obtained when using $\alpha=50$, as well as PSNR scores reported in decibels (dB).

\begin{table}[H]
    \centering
    \begin{tabular}{|c|c|c|cc|c|}\toprule
~~\textbf{dim}~~&~~\textbf{FID - MM-10 $\downarrow$}~~&~~\textbf{FID - MM-\#dim $\downarrow$}~~&\multicolumn{2}{|c|}{~~\textbf{FID - PMF $\downarrow$}~~}&~~\textbf{PSNR $\uparrow$}~~\\\midrule
        8   & 19.66 & 19.46 & 16.88 & $k=2\phantom{0}$ & 15.35 \\
        16  & 15.90 & 16.25 & 13.24 & $k=2\phantom{0}$ & 16.89 \\
        32  & 14.18 & 14.43 & 12.99 & $k=4\phantom{0}$ & 18.30 \\
        64  & \textcolor{blue}{13.24} & 13.60 & 12.65 & $k=6\phantom{0}$ & 19.62 \\
        128 & \textcolor{red}{12.98} & \textcolor{blue}{13.57} & \textcolor{blue}{12.42} & $k=8\phantom{0}$ & \textcolor{blue}{20.81} \\
        256 & 13.31 & \textcolor{red}{13.47} & \textcolor{red}{11.90} & $k=20$ & \textcolor{red}{21.91} \\\bottomrule
    \end{tabular}
    \caption{FID and PSNR results on the Celeba dataset. FID-MM-10 and FID-MM-\#dim denote logistic normal mixture sampling with a number of components equal to ten and the latent space dimension respectively. FID-Mass denotes probability mass function sampling, where the number of bins $k$ is indicated for each model. \textcolor{red}{Red} values represent the best and \textcolor{blue}{blue} the second best.}
    \label{tab:celeba_fid_results}
\end{table}

Table \ref{tab:celeba_fid_results} illustrates that the Simplex AE model is able to achieve competitive FID and PSNR scores on the high-resolution Celeba dataset, despite the restriction of the latent space to a simplex. The effectiveness of selecting the number of components in a mixture model as a heuristic is further supported by the comparable FID scores obtained using either ten components or the full latent dimension. Furthermore, comparing Table \ref{tab:celeba_fid_results} to Table \ref{tab:cifar_fid_results} highlights that PMF sampling consistently results in significantly better FID scores.

Figure \ref{fig:k_study_celeba} demonstrates how the FID score varies with the number of partitions $k$ in PMF sampling on the Celeba dataset. As we can see, when the dimension of the latent space is low, the value of $k$ has little impact on the FID scores and increasing it may even lead to worse results. However, when the dimension is $\dim=64$ or higher, we observe an improvement in FID scores as $k$ increases. This can be attributed to the fact that with a low-dimensional latent space, the model is not able to fully capture all of the information present in the input images.

\begin{figure}[H]
    \centering
    \includegraphics[width=.8\textwidth]{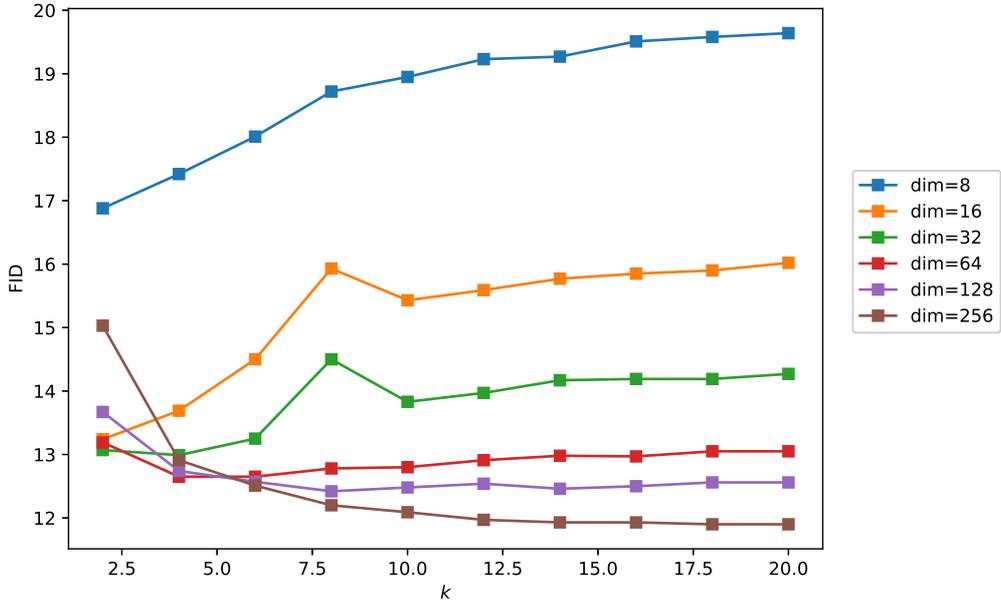}
    \caption{Evolution of the FID as a function of $k$ in PMF sampling on the Celeba dataset.}
    \label{fig:k_study_celeba}
\end{figure}

\subsubsection{Comparison with state-of-the-art methods}
In this section, our goal is to provide a general idea of the FID results that can be expected when using the Simplex AE model. We do not aim to (and we do not, barring for one dataset) directly break the FID records on the datasets used, as those records are systematically held by GANs. We do however demonstrate that Simplex AEs largely improve over the other \textsl{AE strategies} for which results have been published so far. Note that for MNIST we do improve the current GAN-held FID record by $0.21$, which is quite surprising for an AE. 
We also note that this is the first iteration of the Simplex AE model and multiple implementations and future preprocessing choices might further improve the final FID score.

Table \ref{tab:stateofart} summarizes the best FID results we obtained using the Simplex AE model compared to models from the literature. The FID scores of previous methods are reported from the review work of Chadebec et al. \cite{pythae} and Nakagawa et al. \cite{gwae}. We use the results from \cite{pythae} and \cite{gwae} because they use an Autoencoder neural network architecture similar to ours and report results using Gaussian mixture models in the case of \cite{pythae}. The Simplex AE results are taken from Table \ref{tab:mnist_fid_results} and \ref{tab:cifar_fid_results} except for Celeba as the images used in the reported previous work use images of size $64\times64$. Consequently, to obtain comparable results, for Celeba we took the hyperparameter of the best model on this dataset from Table \ref{tab:celeba_fid_results} and ran experiments using images interpolated to $64\times64$.

\begin{table}[H]
    \centering
    \begin{tabular}{|c|cc|cc|cc|}\toprule
~~\textbf{Model}~~                       &\multicolumn{2}{|c|}{~~\textbf{MNIST}~~}                      & \multicolumn{2}{|c|}{~~\textbf{CIFAR-10}~~}                   & \multicolumn{2}{|c|}{~~\textbf{Celeba}~~}\\\midrule
~~AE~~                                   &\multicolumn{2}{|c|}{~~\phantom{2}9.3~~}                      & \multicolumn{2}{|c|}{~~\phantom{1}97.3~~}                     & \multicolumn{2}{|c|}{~~55.4~~} \\
~~VAE \cite{VAE}~~                       &\multicolumn{2}{|c|}{~~26.9~~}                                & \multicolumn{2}{|c|}{~~235.9~~}                               & \multicolumn{2}{|c|}{~~52.4~~} \\
~~$\beta$-VAE \cite{beta-vae}~~          &\multicolumn{2}{|c|}{~~\phantom{2}9.2~~}                      & \multicolumn{2}{|c|}{~~\phantom{1}92.2~~}                     & \multicolumn{2}{|c|}{~~51.7~~} \\
~~WAE \cite{wae}~~                       &\multicolumn{2}{|c|}{~~\phantom{2}8.6~~}                      & \multicolumn{2}{|c|}{~~\phantom{1}96.5~~}                     & \multicolumn{2}{|c|}{~~51.6~~} \\
~~GWAE \cite{gwae}~~                     &\multicolumn{2}{|c|}{~~14.4~~}                                & \multicolumn{2}{|c|}{~~59.9~~}                                & \multicolumn{2}{|c|}{~~45.3~~} \\
~~RAE-L2 \cite{RAE}~~                    &\multicolumn{2}{|c|}{~~9.1~~}                                 & \multicolumn{2}{|c|}{~~85.3~~}                                & \multicolumn{2}{|c|}{~~55.2~~} \\
~~VAE-GAN \cite{vaegan}~~                &\multicolumn{2}{|c|}{~~\phantom{2}6.3~~}                      & \multicolumn{2}{|c|}{~~197.5~~}                               & \multicolumn{2}{|c|}{~~35.6~~} \\ \midrule
~~$\mbox{Simplex AE}_{\best}$ MM-10~~    &\multicolumn{2}{|c|}{~~\phantom{2}5.75~~}                     & \multicolumn{2}{|c|}{~~\phantom{1}15.35~~}                    & \multicolumn{2}{|c|}{~~\textcolor{blue}{10.65}~~}    \\ \midrule
~~$\mbox{Simplex AE}_{\best}$ MM-\#dim~~ &\multicolumn{2}{|c|}{~~\textcolor{blue}{\phantom{2}5.21}~~}   & \multicolumn{2}{|c|}{~~\textcolor{blue}{\phantom{1}14.90}~~}  & \multicolumn{2}{|c|}{~~10.67~~} \\ \midrule
~~$\mbox{Simplex AE}_{\best}$ PMF~~      & \textcolor{red}{\phantom{2}4.29} &$k=10$                     & \textcolor{red}{\phantom{1}13.55} & $k=2$                     & \textcolor{red}{9.26} & $k=20$ \\ \bottomrule
    \end{tabular}
    \caption{FID comparison of Simplex AE with other Autoencoder based methods. $\mbox{Simplex AE}_{\best}$ denotes the best FID score obtained with a Simplex AE. \textcolor{red}{Red} values represent the best and \textcolor{blue}{blue} the second best.} 
    \label{tab:stateofart}
\end{table}

The results in Table \ref{tab:stateofart} show that the Simplex AE with Mixture model sampling using ten components outperforms other AEs on all tested datasets. We stress again that we do not outperform GAN-held records, except for the MNIST dataset. This suggests that limiting the latent space to a simplex does not negatively impact the FID score. Additionally, the proposed heuristic for determining the number of components in a mixture model was found to improve the FID score on all datasets compared to using the number of classes. Using PMF sampling instead of MM-10 sampling further improved the FID score by $1.46$, $1.8$, and $1.39$ on the MNIST, CIFAR-10, and Celeba datasets, respectively. 

The best AE FID results to date ont the MNIST, CIFAR-10, and Celeba datasets are respectively $6.3$, $85.3$ and $35.6$ we hence substantially improve those figures. As we underlined, AEs are not the best performing algorithms on the concerned datasets and all records are currently held by GANs. While we do not perform better than GANs on CIFAR and Celeba we do manage to squeeze-out a non-negligible improvement (of $0.21$) and break the current GAN-held record for the MNIST dataset.


\section{Conclusion and future work}
In this work, we address the question of "How can we improve the sampling from the latent space of an Autoencoder?" To address this question, we first model the latent space as a simplex, which imposes boundaries on the space and makes sampling easier due to its finite nature. We then introduce a novel mixture model sampling formulation based on logistic normal distributions, which allows for the sampling of points on a simplex. Additionally, we develop a heuristic for determining the number of components in the mixture model based on the number of vertices in the simplex, which is independent of the number of classes. Finally, we propose a sampling method based on probability mass functions. Our experiments demonstrate that restricting the latent space does not negatively impact the FID performance, and the proposed sampling methods result in non-negligible performance gains in terms of FID score on multiple datasets. The Simplex AE model achieves an image generation FID of 4.29, 13.55, and 11.90 on the MNIST, CIFAR-10, and Celeba datasets, respectively.

Future extensions to this work may include using the probability mass function to interpolate between two points in the latent space via high density areas, or finding geodesics. One may also explore improving the robustness of the method to adversarial attacks. Those improvements are left as further research that can build upon the current work.

\section*{Acknowledgement}
The authors were granted access to the HPC resources of MesoPSL financed by the Région Île-de-France and the Equip@Meso project (reference ANR-10-EQPX-29-01) of the \textsl{programme investissements d’avenir} supervised by France's \textsl{Agence nationale pour la recherche}.

\bibliographystyle{apalike}

\appendix

\newpage
\section{Datasets}
\subsection{MNIST}
The MNIST dataset, introduced in \cite{mnist}, consists of ten classes of grayscale images, each with a size of $28\times 28$. The dataset is divided into three sets: a training set with $50,000$ images, a validation set with $10,000$ images, and a test set with $10,000$ images. In our experiments, we interpolate the images to $64\times 64$ and duplicate them along the channels axis to create three channels.

\subsection{CIFAR-10}
The CIFAR-10 dataset, introduced in \cite{cifar10}, is a collection of $32\times 32$ color images of real-world objects such as airplanes and boats. It contains a total of $60,000$ images, which are divided into ten classes. For our experiments, we split the dataset into three sets: a training set with $40,000$ images, a validation set with $10,000$ images, and a test set with $10,000$ images. We also interpolate the images to a size of $64\times 64$.

\subsection{Celeba}
The Celebrity Faces dataset, introduced in \cite{celeba}, consists of color images of celebrity faces with a size of $178\times 218$. It includes a total of $200,000$ images. For our experiments, we use the original train, validation, and test dataset split provided with the dataset.

\section{Network architectures}
\subsection{For the synthetic dataset}
\label{sythetic_dataset_ae_archi}

\begin{table}[H]
    \centering
    \begin{tabular}{|c|c|c|}
        \toprule
        ~~\textbf{Layers}~~ & ~~\textbf{Encoder}~~ & ~~\textbf{Decoder}~~ \\
         & ~~$\mbox{Input size}=[20]$~~& ~~$\mbox{Input size}=[\#\dim]$~~\\\hline
        \multirow{2}{*}{Layer 1} & $\lin(20, 10)$& $\lin(\#\dim,5)$ \\
         & ReLU & ReLU \\\hline
        \multirow{2}{*}{Layer 2} &  $\lin(10,\phantom{1}5)$ &  $\lin(\phantom{1}5,10)$\\
         & ReLU & ReLU \\\hline
        \multirow{2}{*}{Layer 3} &  $\lin(5,\#\dim)$ &  $\lin(10, 20)$\\
         & Softmax & \\
        \bottomrule
    \end{tabular}
    \caption{Autoencoder neural network architecture employed for the synthetic dataset.}
\end{table}

\subsection{For benchmark datasets}
\label{benchmark_dataset_ae_archi}

\begin{table}[H]
    \centering
    \begin{tabular}{|c|c|c|}
        \toprule
        ~~\textbf{Layers}~~ & ~~\textbf{Encoder}~~ & ~~\textbf{Decoder}~~ \\
         & ~~$\mbox{Input size}=[3, 64, 64]$ ~~& ~~$\mbox{Input size}=[\#\dim]$ ~~\\\hline
        \multirow{2}{*}{Layer 1} & $\conv(\phantom{1}32,(4,4),s=2,p=1)$ & $\lin(\#\dim,256)$ \\
         & SiLU & SiLU \\
         \hline
        \multirow{3}{*}{Layer 2} & $\conv(\phantom{1}64,(4,4),s=2,p=1)$ & $\lin(256,256\times 4 \times 4)$ \\
         & SiLU & SiLU \\
         &  & $\resh(256,4,4)$\\
        \hline
        \multirow{2}{*}{Layer 3} & $\conv(128, (4,4), s=2, p=1)$ & $\convT(128,(4,4),s=2,p=1)$ \\
         & SiLU & SiLU \\
        \hline
        \multirow{3}{*}{Layer 4} & $\conv(256, (4,4), s=2, p=1)$ & $\convT(\phantom{1}64,(4,4),s=2,p=1)$ \\
         & SiLU & SiLU\\
         & $\resh(256\times 4 \times 4)$ & \\
         \hline
         \multirow{2}{*}{Layer 5} & $\lin(256 \times 4 \times 4,\#\dim)$ & $\convT(\phantom{1}32,(4,4),s=2,p=1)$ \\
         & Softmax & SiLU\\\hline
         \multirow{2}{*}{Layer 6} &- & $\convT(\phantom{12}3,(4,4),s=2, p=1)$ \\
         &  & Sigmoid \\
         
        \bottomrule
    \end{tabular}
    \caption{Autoencoder neural network architecture employed to benchmark the MNIST and CIFAR-10 datasets. $s$ denotes the stride and $p$ denotes padding.}
\end{table}

\begin{table}[H]
    \centering
    \begin{tabular}{|c|c|c|}
        \toprule
        ~~\textbf{Layers}~~ & ~~\textbf{Encoder}~~ & ~~\textbf{Decoder}~~ \\
         & ~~$\mbox{Input size}=[3, 178, 218]$ ~~& ~~$\mbox{Input size}=[\#\dim]$ ~~\\\hline
        \multirow{2}{*}{Layer 1} & $\conv(\phantom{1}32,(4,4),s=2,p=1)$ & $\lin(\#\dim,256)$ \\
         & SiLU & SiLU \\
         \hline
        \multirow{3}{*}{Layer 2} & $\conv(\phantom{1}64,(4,4),s=2,p=1)$ & $\lin(256,256\times 13 \times 11)$ \\
         & SiLU & SiLU \\
         &  & $\resh(256,13,11)$\\
        \hline
        \multirow{2}{*}{Layer 3} & $\conv(128, (4,4), s=2, p=1)$ & $\convT(128,(3,4),s=2,p=(0,1)$ \\
         & SiLU & SiLU \\
        \hline
        \multirow{3}{*}{Layer 4} & $\conv(256, (4,4), s=2, p=1)$ & $\convT(\phantom{1}64,(3,4),s=2,p=(0,1))$ \\
         & SiLU & SiLU\\
         & $\resh(256\times 13 \times 11)$ & \\
         \hline
         \multirow{2}{*}{Layer 5} & $\lin(256 \times 13 \times 11,\#\dim)$ & $\convT(\phantom{1}32,(3,4),s=2,p=(1,1))$ \\
         & Softmax & SiLU\\\hline
         \multirow{2}{*}{Layer 6} &- & $\convT(\phantom{12}3,(2,4),s=2, p=(0,0))$ \\
         &  & Sigmoid \\
         
        \bottomrule
    \end{tabular}
    \caption{Autoencoder neural network architecture employed to benchmark the Celeba datasets. $s$ denotes the stride and $p$ denotes padding.}
\end{table}

\section{Synthetic image samples}
\label{img_samples}

\subsection*{Synthetic MNIST image samples}
\begin{figure}[H]
    \centering
     \begin{subfigure}[b]{\textwidth}
         \centering
         \includegraphics[width=\textwidth]{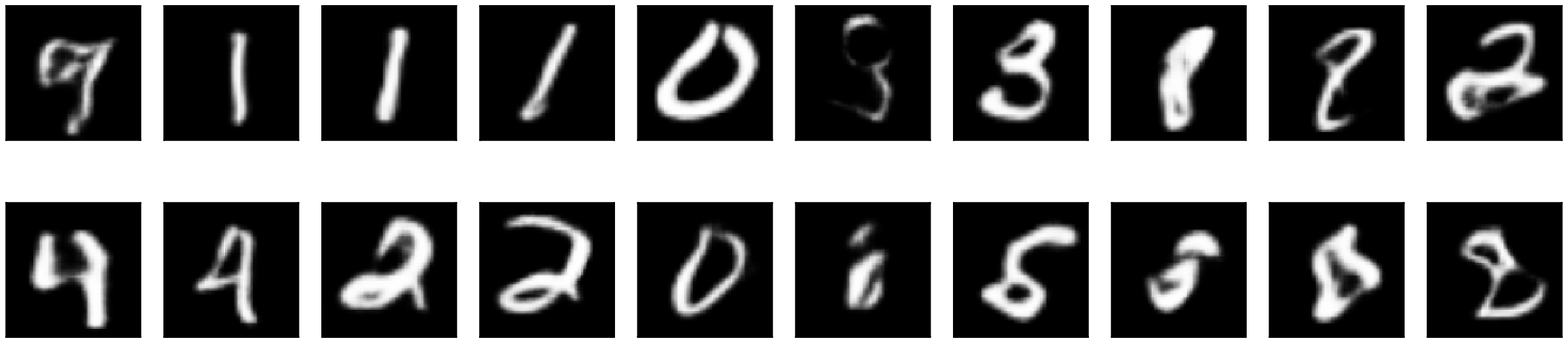}
         \caption{$\mbox{Simplex AE}_{32}$ MM-10 sampling.}
     \end{subfigure}
     \hfill
     \begin{subfigure}{\textwidth}
         \centering
         \includegraphics[width=\textwidth]{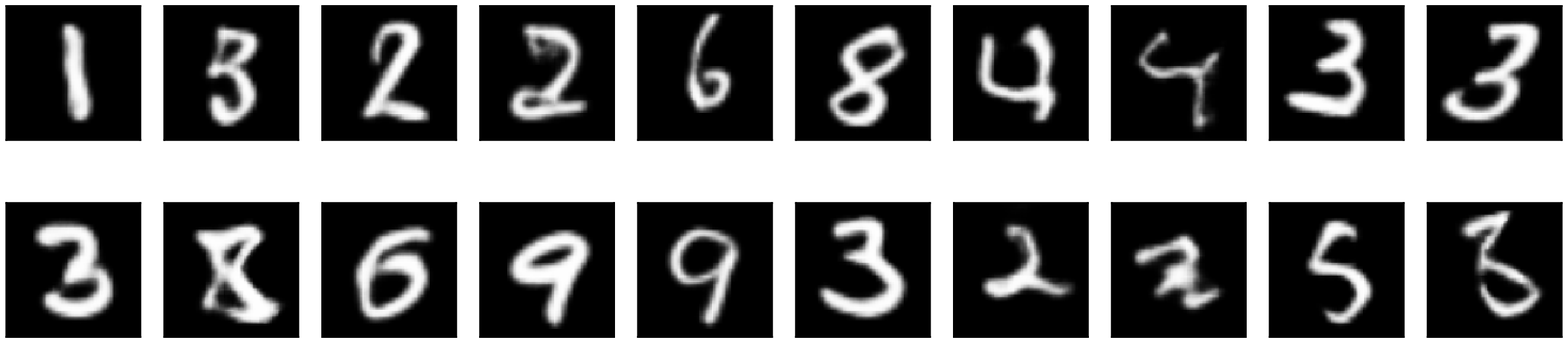}
         \caption{$\mbox{Simplex AE}_{32}$ MM-$\dim$ sampling.}
     \end{subfigure}
     \hfill
     \begin{subfigure}[b]{\textwidth}
         \centering
         \includegraphics[width=\textwidth]{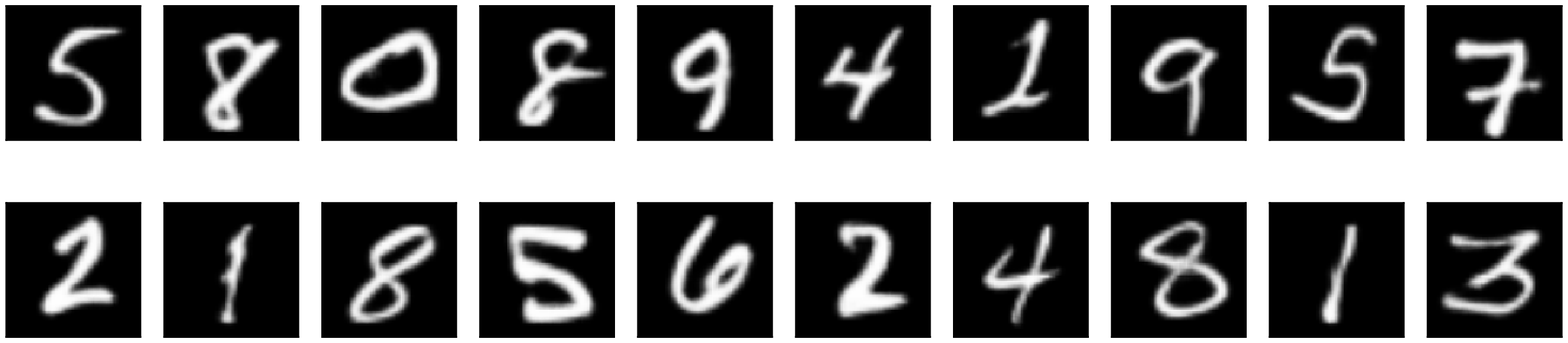}
         \caption{$\mbox{Simplex AE}_{32}$ PMF sampling $k=10$.}
     \end{subfigure}

    \caption{Synthetic images from the MNIST dataset with Simplex AE using Mixture model sampling, and probability mass function sampling. No cherry-picking was done to select the images.}
    \label{fig:mnist_sample_images}
\end{figure}

\subsection*{Synthetic CIFAR-10 image samples}
\begin{figure}[H]
    \centering
     \begin{subfigure}[b]{\textwidth}
         \centering
         \includegraphics[width=\textwidth]{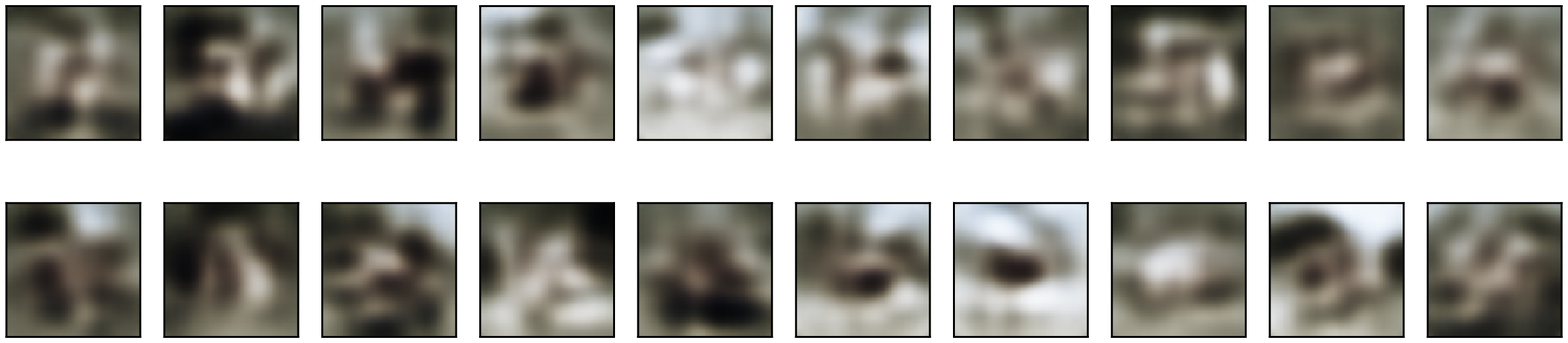}
         \caption{$\mbox{Simplex AE}_{64}$ MM-10 sampling .}
     \end{subfigure}
     \hfill
     \begin{subfigure}{\textwidth}
         \centering
         \includegraphics[width=\textwidth]{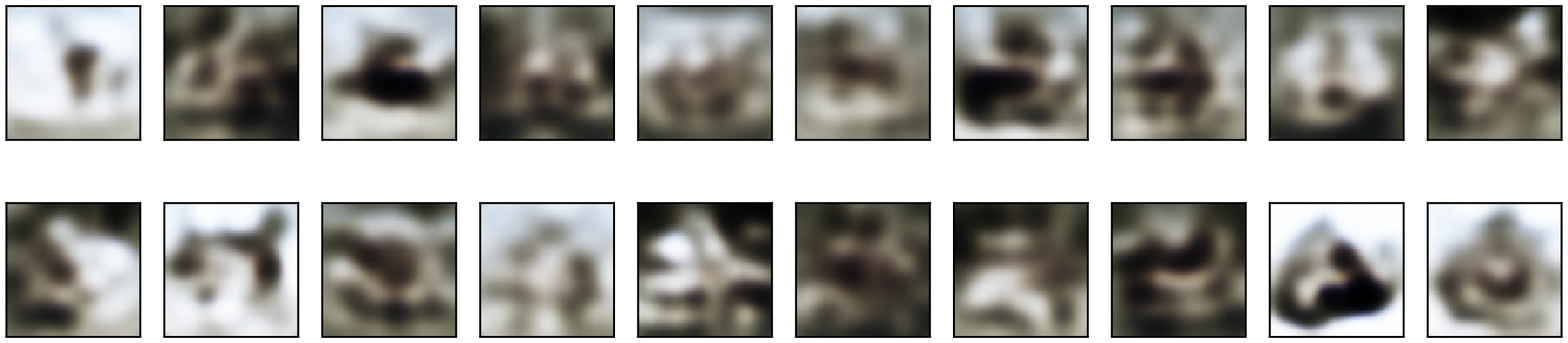}
         \caption{$\mbox{Simplex AE}_{256}$ MM-$\dim$ sampling.}
     \end{subfigure}
     \hfill
     \begin{subfigure}[b]{\textwidth}
         \centering
         \includegraphics[width=\textwidth]{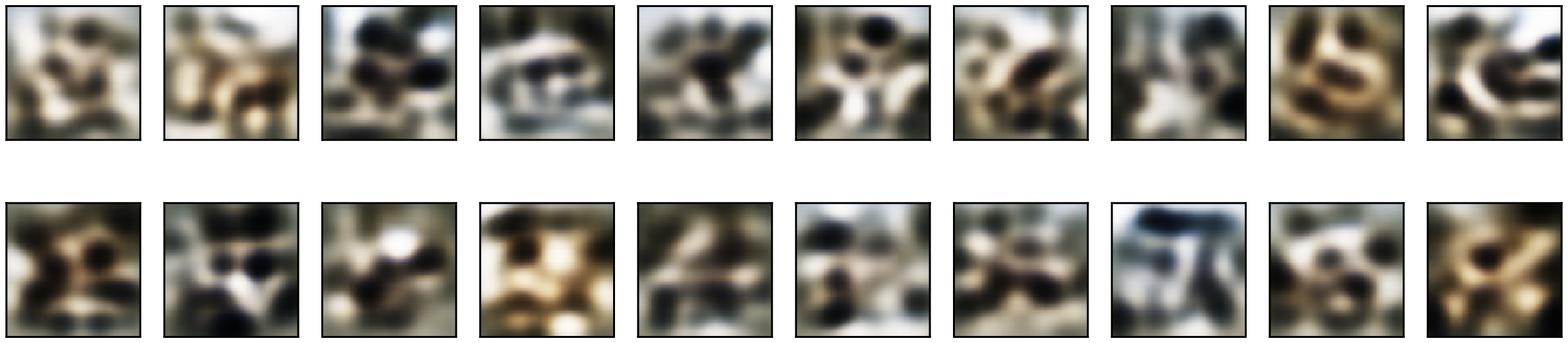}
         \caption{$\mbox{Simplex AE}_{128}$ PMF sampling $k=2$.}
     \end{subfigure}
     \hfill
     \begin{subfigure}[b]{\textwidth}
         \centering
         \includegraphics[width=\textwidth]{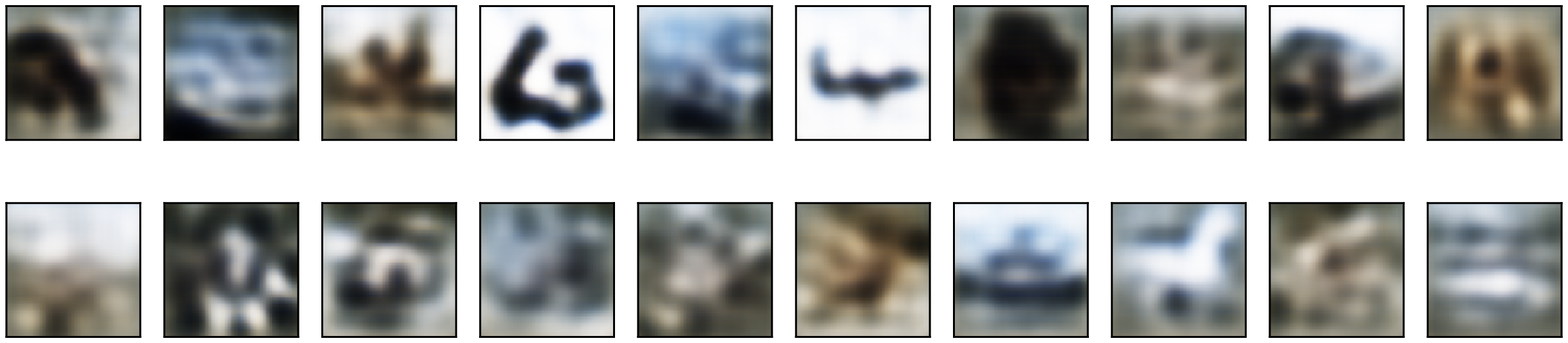}
         \caption{$\mbox{Simplex AE}_{128}$ PMF sampling $k=20$.}
     \end{subfigure}

    \caption{Synthetic images from the CIFAR-10 dataset with Simplex AE using Mixture model sampling, and probability mass function sampling. No cherry-picking was done to select the images.}
    \label{fig:cifar_sample_images}
\end{figure}

\subsection*{Synthetic Celeba image samples}
\begin{figure}[H]
    \centering
     \begin{subfigure}[b]{\textwidth}
         \centering
         \includegraphics[width=\textwidth]{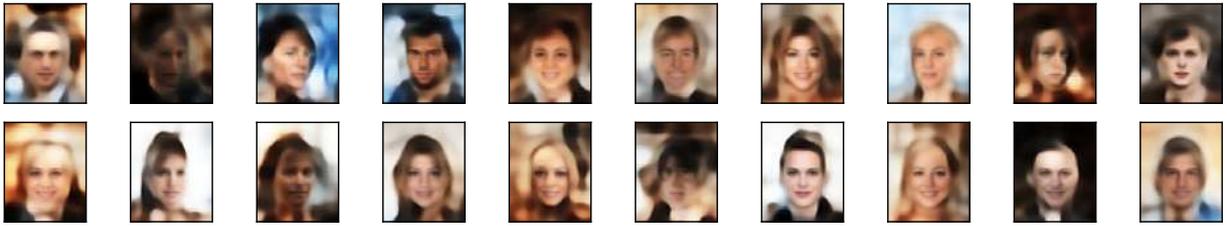}
         \caption{$\mbox{Simplex AE}_{128}$ MM-10 sampling.}
     \end{subfigure}
     \hfill
     \begin{subfigure}{\textwidth}
         \centering
         \includegraphics[width=\textwidth]{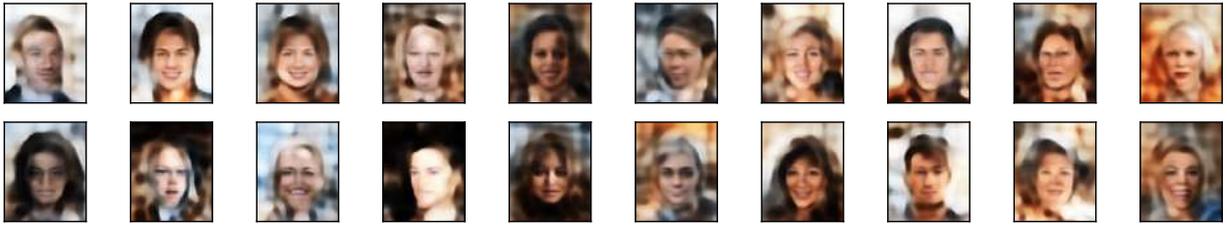}
         \caption{$\mbox{Simplex AE}_{256}$ MM-$\dim$ sampling.}
     \end{subfigure}
     \hfill
     \begin{subfigure}[b]{\textwidth}
         \centering
         \includegraphics[width=\textwidth]{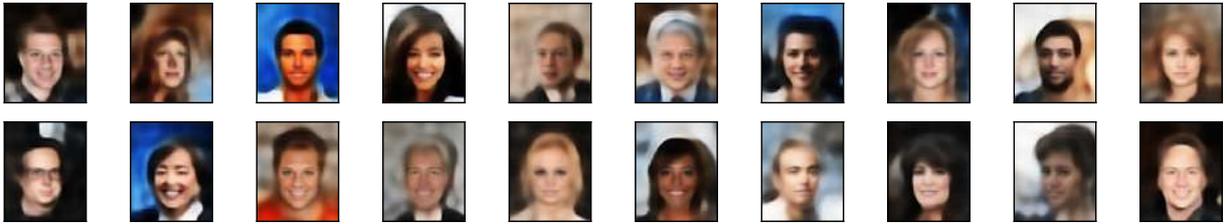}
         \caption{$\mbox{Simplex AE}_{256}$ PMF sampling $k=20$.}
     \end{subfigure}

    \caption{Synthetic images from the Celeba dataset with Simplex AE using Mixture model sampling, and probability mass function sampling. No cherry-picking was done to select the images.}
    \label{fig:celeba_sample_images}
\end{figure}

\newpage
\section{Parameter \texorpdfstring{$\alpha$}{} study}
\label{alpha_study}
Table \ref{alpha_mnist} and \ref{alpha_cifar} demonstrate that the value of $\alpha$ has a minimal effect on the FID score, regardless of the sampling strategy. Therefore, in our experiments, we selected $\alpha=30$ for both the MNIST and CIFAR-10 datasets. For the MNIST dataset, $\alpha=30$ is the value that precedes a sudden increase in the FID using the MM-10 sampling method. For the CIFAR-10 dataset, we chose the same value for consistency, as no specific value stands out in our study. For the Celeba datatset we chose a value of $\alpha=50$ as it gives a relatively lower FID value.

\begin{table}[H]
    \centering
    \begin{tabular}{|c|c|c|c|}\toprule
~~\textbf{$\alpha$}~~&~~\textbf{FID - MM-10 $\downarrow$}~~&~~\textbf{FID - MM-16 $\downarrow$}~~ & ~~\textbf{KNN accuracy $\uparrow$}~~\\\midrule
        0.1 & 6.22 & \textcolor{blue}{6.01} & 94.62\% \\
        0.5 & 6.05 & 6.02 & 96.03\%  \\
        1 & 6.09 & \textcolor{red}{6.00} & 95.94\%   \\
        10 & \textcolor{blue}{6.05} & 6.05 & \textcolor{blue}{96.25\%}  \\
        30 & \textcolor{red}{6.03} & 6.07 & \textcolor{red}{96.44\%}  \\
        50 & 6.20 & 6.02 & 96.12\%  \\\bottomrule
    \end{tabular}
    \caption{FID and KNN classification results on the MNIST dataset. FID-MM-10 and FID-MM-16 denote logistic normal mixture sampling with a number of components equal to $10$ and $16$ respectively. \textcolor{red}{Red} values represent the best and \textcolor{blue}{blue} the second best.}
    \label{alpha_mnist}
\end{table}

\begin{table}[H]
    \centering
    \begin{tabular}{|c|c|c|c|}\toprule
~~\textbf{$\alpha$}~~&~~\textbf{FID - MM-10 $\downarrow$}~~&~~\textbf{FID - MM-32 $\downarrow$}~~ & ~~\textbf{KNN accuracy $\uparrow$}~~\\\midrule
        0.1 & 15.62 & 15.17 & 27.89\% \\
        0.5 & 15.76 & 15.78 & 27.50\%  \\
        1 & 15.55 & 15.53 & 27.42\% \\
        5 & 15.74 & 15.64 & 27.16\%\\ 
        10 & \textcolor{blue}{15.45} & 15.48 & 27.78\% \\
        20 & 15.58 & 15.56 & 27.61\%\\ 
        30 & 15.79 & 15.68 & 27.71\%  \\
        40 & 15.71 & 15.62 & 27.63\%  \\
        50 & \textcolor{red}{15.20} & \textcolor{red}{15.28} & \textcolor{red}{28.72\%} \\
        60 & 15.46 & \textcolor{blue}{15.46} & 27.24\%\\
        70 & 15.73 & 15.69 & 27.70\%\\
        80 & 15.92 & 15.81 & 27.66\%\\
        90 & 15.61 & 15.61 & \textcolor{blue}{28.01\%}\\
        100 & 15.82 & 15.83 & 27.67\%\\ \bottomrule
    \end{tabular}
    \caption{FID and KNN classification results on the CIFAR-10 dataset. FID-MM-10 and FID-MM-32 denote logistic normal mixture sampling with a number of components equal to $10$ and $32$ respectively. \textcolor{red}{Red} values represent the best and \textcolor{blue}{blue} the second best.}
    \label{alpha_cifar}
\end{table}

\begin{table}[H]
    \centering
    \begin{tabular}{|c|c|c|}\toprule
~~\textbf{$\alpha$}~~&~~\textbf{FID - MM-10 $\downarrow$}~~&~~\textbf{FID - MM-64 $\downarrow$}~~\\\midrule
        0.1 & 13.59 & 13.94  \\
        0.5 & 13.48 & 13.94   \\
        1 & 13.50 & 13.84 \\
        10 & 13.54 & 13.76 \\
        20 & 13.61 & 14.01 \\ 
        30 & 13.78 & 14.08 \\
        40 & 13.54 & 13.92 \\
        50 & \textcolor{red}{13.18} & \textcolor{red}{13.60}  \\
        60 & 13.56 & 13.82 \\
        70 & 13.53 & 13.89 \\
        80 & 13.47 & 13.78 \\
        90 & 13.40 & 13.75 \\
        100 & \textcolor{blue}{13.29} & \textcolor{blue}{13.69} \\ \bottomrule
    \end{tabular}
    \caption{FID results on the Celeba dataset. FID-MM-10 and FID-MM-64 denote logistic normal mixture sampling with a number of components equal to $10$ and $64$ respectively. \textcolor{red}{Red} values represent the best and \textcolor{blue}{blue} the second best.}
    \label{alpha_celeba}
\end{table}

\end{document}